\documentclass[a4paper,12pt]{elsarticle}
\usepackage{hyperref}
\usepackage{mathtools} 
\usepackage{booktabs} 
\usepackage{tikz} 
\usepackage{color}
\usepackage{etoolbox}
\usepackage{appendix}
\usepackage{bbm}
\usepackage{enumitem}
\usepackage{amsmath}
\usepackage{amssymb}
\usepackage{comment}
\usepackage{geometry}
\usepackage{setspace}
\usepackage{graphicx}	
\usepackage{algorithm}
\usepackage{multicol}
\usepackage{subcaption}
\usepackage{algpseudocode}
\usepackage{amsthm}

\usepackage{xcolor}

\newtheorem{innercustomgeneric}{\customgenericname}
\providecommand{\customgenericname}{}
\newcommand{\newcustomtheorem}[2]{%
  \newenvironment{#1}[1]
  {%
   \renewcommand\customgenericname{#2}%
   \renewcommand\theinnercustomgeneric{##1}%
   \innercustomgeneric
  }
  {\endinnercustomgeneric}
}

\newcustomtheorem{customthm}{Theorem}
\newcustomtheorem{customlemma}{Lemma}
\newcustomtheorem{customcorollary}{Corollary}

\makeatletter
\def\ps@pprintTitle{%
 \let\@oddhead\@empty
 \let\@evenhead\@empty
 \def\@oddfoot{}%
 \let\@evenfoot\@oddfoot}
\makeatother
\biboptions{authoryear}

\begin{document}

\begin{frontmatter}

\author[add1]{Saketh Reddy Karra}
\ead{skarra7@uic.edu}
\author[add1]{Son The Nguyen}
\ead{snguye65@uic.edu}
\author[add1]{Theja Tulabandhula}
\ead{theja@uic.edu}

\address[add1]{University of Illinois Chicago, Chicago, United States}

\title{Estimating the Personality of White-Box Language Models}

 \begin{abstract}
   Technology for open-ended language generation, a key application of artificial intelligence, has advanced to a great extent in recent years. Large-scale language models, which are trained on large corpora of text, are being used in a wide range of applications everywhere, from virtual assistants to conversational bots. While these language models output fluent text, existing research shows that these models can and do capture human biases. Many of these biases, especially those that could potentially cause harm, are being well-investigated. On the other hand, studies that infer and change human personality traits inherited by these models have been scarce or non-existent. Our work seeks to address this gap by exploring the personality traits of several large-scale language models designed for open-ended text generation and the datasets used for training them. We build on the popular Big Five factors and develop robust methods that quantify the personality traits of these models and their underlying datasets. In particular, we trigger the models with a questionnaire designed for personality assessment and subsequently classify the text responses into quantifiable traits using a Zero-shot classifier. Our estimation scheme sheds light on an important anthropomorphic element found in such AI models and can help stakeholders decide how they should be applied as well as how society could perceive them. Additionally, we examined approaches to alter these personalities, adding to our understanding of how AI models can be adapted to specific contexts.
 \end{abstract}


\end{frontmatter}

\section{Introduction}
With advancements in methodologies and increasing availability of computational resources for training deep neural networks in recent years, Natural Language Processing (NLP) research has seen substantial progress on a variety of tasks such as language modeling~\citep{liu2018learning}, question answering~\citep{soares2020literature}, and machine translation~\citep{zong2018application} to name a few. In particular, Natural Language Generation (NLG) models that enable the generation of human-readable language have become the central building blocks of modern artificial intelligence (AI) applications, such as virtual assistants, chatbots, automatic translators, and text summarizers~\citep{bai2022training}.

Incorporating personality into AI applications has gained increasing attention in recent years, as seen from their use in finance~\citep{spencer2018designing} and e-commerce sectors ~\citep{ahmad2020introducing}. Personality is a complex construct that encompasses a range of behaviors, emotional states, and thought patterns arising from biological and environmental factors. In the context of AI, personality traits can be thought of as the manifestation of human-like characteristics in language models, which are typically trained on human-generated text from the Internet. Consequently, these models indirectly incorporate human tendencies and biases reflected in the written text unless data pre-processing steps are taken. Recent studies ~\citep{bordia2019identifying,sheng2019woman} have investigated the incorporation of such biases into language models, highlighting the need for careful consideration of personality traits in the development of AI systems. In this direction, Our paper is among the first to validate and quantify the personality traits of language models empirically. We explore ways to modify these traits to meet specific requirements, thereby enhancing the effectiveness and ethical considerations of AI applications in various contexts. 

Identifying personality traits in language models can have significant implications for their design, implementation, and deployment. In terms of user experience, understanding the personality of a language model can lead to creating more engaging and empathetic models, which can better interact with users and improve user satisfaction, ultimately increasing user adoption of AI-based applications. Moreover, by identifying personality traits, researchers can uncover underlying biases in language models, leading to bias reduction and improving the fairness and accuracy of AI systems. Additionally, identifying personality traits can help identify the most appropriate use cases for specific language models. For example, a language model with a friendly personality may be better suited for customer service applications. In contrast, a more formal personality may be more appropriate for professional training applications. These potential implications highlight the importance of considering the personality traits of language models in their development and deployment. Moreover, we can aim to design learning systems that enable language models to alter their personality traits over time in response to changing environmental factors. This would result in dynamic AI systems that are responsive and more engaging to users over a long time frame. Extending this research could lead to further improvements in the development of chatbots, virtual assistants, and other AI systems, enabling them to provide a more human-like experience to users.

Our research focuses on quantifying the personality traits of white-box language models and exploring how to change them.  Investigating the personality traits of language models is a challenging task, as the models' outputs are stochastic and require a suitable auxiliary verification model to elicit traits from the text responses (for example, manual human verification may not be feasible).  For modifying personalities, one direct approach is to train or fine-tune language models on personality-annotated text data, taking inspiration from human psychological research. We show that this too is challenging, especially if we are interested in achieving precise control over personality changes. In this paper, we aim to address these challenges and investigate the personality traits of popular language models in open-ended text generation, along with the datasets used to train them. Our key hypothesis is that language models generate text responses reflecting the personality traits of the datasets on which they were trained when prompted. To quantify their personalities, we process the datasets as well as the text responses generated by the language models using state-of-the-art auxiliary prediction models. However, when it comes to altering the persona of language models, we limit ourselves to evaluating white-box language models that allow users to access and modify parameters, unlike black-box models like ChatGPT~\citep{ray2023chatgpt}. Additionally, we focus our attention solely on foundation models and do not consider instruction-tuned models. Unlike foundation models, instruction-tuned models use supervised instruction fine-tuning to understand the tasks and do not require prompt engineering. Furthermore, evaluating the intrinsic personalities of instruction-tuned models can be challenging since they can be trained to embody any desired personality in their responses successfully.

To the best of our knowledge, we are one of the first to explore personality traits of large language models in general, complementing the literature that studies biases. To recap, the primary contributions of this work are: 
\begin{itemize}
    \item We explore the \emph{five-factor method}~\citep{digman1990personality}, a widely used personality assessment questionnaire, to quantify the personality traits of datasets and language models.
    
    \item We propose a novel estimation scheme for evaluating the personality traits of both the datasets and the language models, using Zero-shot learning (ZSL)~\citep{yin2019benchmarking}. Our approach involves triggering language models with the Big Five questionnaire and subsequently classifying the resulting text responses into quantifiable personality traits.
    
    \item Finally, we discuss approaches to alter the personalities of these models by finetuning as well as by using a pretrained auxiliary personality classification model.
    
    \item  We perform extensive experiments on multiple white-box language models and their underlying training corpora, explicitly contrasting how the personality profiles differ between the two. The results validate the effectiveness of our proposed methodology and give insights into what factors lead to differences between models and data, as well as across models.
\end{itemize}

The rest of the paper is structured as follows. In Section \ref{sec:rw}, we discuss closely related work. In Section \ref{sec:preliminaries}, we introduce the Big Five factors, the Zero-shot classifier, and the pretrained language models. In Section \ref{sec:Methods}, we propose the methods to evaluate the personality traits of datasets and language models, followed by a discussion on approaches to alter these traits. In Section \ref{sec:experiments}, we discuss the experimental setup and extensively document the traits of datasets and the corresponding language models. In addition, we also provide preliminary results for altering these personalities. We conclude with some comments on future work in Section \ref{sec:conclusion}.

\section{Related Work} \label{sec:rw}
Our work builds on the following streams of research: (a)  measurement of human personality traits, (b) design of personality detection methodologies, and (c) studies on societal biases in open-ended text generation. We briefly discuss some of the works below.

\subsection{Personality Measures}
Researchers have used various schemes for human personality modeling such as 16PF~\citep{schuerger2000sixteen}, EPQ-R~\citep{miles2004eysenck}, Myers–Briggs Type Indicator (MBTI)~\citep{miles2004eysenck}, and the three trait personality models/PEN~\citep{eysenck2012model} among others. For instance, MBTI is a widely adopted personality measure that is sometimes used to screen job candidates. It relies on the theory that random variation in human behavior is quite orderly and consistent due to certain basic differences in the way people prefer to use perception and judgment. The MBTI personality measure categorizes people into two categories in each of the four dimensions: introversion versus extroversion; sensing versus intuiting; thinking versus feeling; and judging versus perceiving. Another popular measure used in the literature on automated personality detection is the Big Five factors measure~\citep{digman1990personality}, which defines and quantifies the following attributes: \emph{Extraversion, Neuroticism, Agreeableness, Conscientiousness, and Openness}. In this work, we adapt and estimate the Big Five factors of white-box language models and their underlying datasets using a novel methodology described in Section~\ref{sec:Methods}.

\subsection{Methods for Automatic Personality Detection}
Automatic detection of personality traits from text produced by humans, instead of them explicitly answering a questionnaire or undergoing a test, has been explored in the literature. \cite{pennebaker1999linguistic} compiled a dataset of anonymous essays tagged with the authors' personalities based on the Big Five factors and used the so-called Linguistic Inquiry and Word Count (LIWC) features to determine the correlation between essays and personality.~\cite{liu2016analyzing} used deep learning-based models in combination with the atomic features of text, i.e., the characters, to predict personality traits of individuals using hierarchical and vectorized word/sentence representations.~\cite{akrami2019automatic} developed a model that can extract Big Five factors from text using machine learning techniques like support vector regression.~\cite{jeremy2019identifying} performed experiments to automatically predict a user's personality based on Big Five factors on Twitter.~\cite{mehta2020recent} provides an overview of state-of-the-art machine learning models for automatic personality detection with a specific focus on multi-modal approaches. 
Recently~\cite{zero-shotclassify} used Zero-shot learning (ZSL) to classify text responses from a self-report questionnaire in terms of the Big Five factors. Through their experiments, the authors showed that a strong positive relationship (e.g., correlation) exists between the ZSL scores and the scores on the self-report questionnaire for each specific trait. While the approaches discussed earlier have shown promising results in prediction tasks, they often rely on large amounts of data for effective performance. To address this challenge, we quantify the traits of our models by using the ZSL framework (see Section~\ref{sec:Methods}). Our novel approach explores all the possible scenarios for defining the personality trait labels and thus robustly adapts the ZSL framework to language models. 

\subsection{Study of Biases in Language Modeling}
Recent works have explored multiple societal biases that are learned by language models, which may sometimes be at odds with the prevailing societal values.~\cite{bolukbasi2016man} demonstrated quantitatively that word embeddings contain biases in their geometry that reflect gender stereotypes present in the broader society.~\cite{sheng2019woman} performed an experiment analyzing different textual contexts where biases can occur for different demographics in text samples generated using the state-of-the-art language models. ~\cite{bordia2019identifying} evaluated the magnitude of gender bias in word-level language models that are trained on WikiText-2 and CNN/Daily Mail datasets.~\cite{nadeem2020stereoset} evaluated popular models such as BERT~\citep{devlin2018bert}, GPT-2~\citep{radford2019language} and XLNet~\citep{yang2019xlnet} using StereoSet dataset and show that these models exhibit strong stereotypical biases in the following four domains: gender, profession, race, and religion. We note that our work is complementary to all these studies in the following sense. While we aim to understand human tendencies captured by language models similar to these prior studies, our narrow but well-defined focus on characterizing the learned personality traits and potentially altering them is a differentiating factor, as well as a first of its kind.

\section{Preliminaries} \label{sec:preliminaries}
In this section, we discuss the Big Five factors and introduce our hypothesis for quantifying these traits in language models and their corresponding training datasets. We also describe the Zero-shot classifier (ZSC) and provide details about the various language models we have analyzed in our study.

\subsection{Big Five factors}\label{sec:bff}
Our study quantifies personality traits using the Big Five factors, also known as the \emph{five-factor model}. Under this model, personality can be reduced to the following five core factors:
\begin{itemize}
    \item \emph{Extraversion}: sociable and energetic versus reserved and solitary.
\item \emph{Neuroticism}: sensitive and nervous versus secure and confident.
\item \emph{Agreeableness}: trustworthy, straightforward, generous, and modest versus unreliable, complicated, meager, and boastful.
\item \emph{Conscientiousness}: efficient and organized versus sloppy and careless.
\item \emph{Openness}: inventive and curious versus dogmatic and cautious.
\end{itemize}
The \emph{neuroticism} factor has a negative connotation in contrast to the other four factors. Therefore, we estimate \emph{Emotional Stability} instead to be consistent with other factors in the rest of the paper.

The questionnaire itself is a list of fifty statements, each referring to different characteristics of an individual. Accordingly, each statement is designed to elicit a specific Big Five factor behavior. In general,  individuals respond to every statement in the questionnaire by opting for one of the following choices: (a) very inaccurate, (b) moderately inaccurate, (c) neither inaccurate nor accurate, (d) moderately accurate, and (e) very accurate. The response to every statement is scored against a predetermined Big Five factor on a scale of $1$ to $5$ as shown in Table \ref{tab:prob-table}. There are ten statements evaluated for the \emph{extraversion} factor, ten questions for \emph{agreeableness}, and so forth. Finally, the aggregated scores against each of the Big Five factors are averaged to obtain the quantifiable trait scores.

\begin{table}[]
\scriptsize
\centering
\begin{tabular}{|c|c|}
\hline
         \textbf{Response}  &   \textbf{Score}         \\ \hline
 Very Inaccurate     & 1            \\ \hline
 Moderately Inaccurate  &  2           \\ \hline
 Neither Inaccurate nor Accurate & 3  \\ \hline
 Moderately Accurate                  &  4            \\ \hline
 Very Accurate                 & 5                 \\ \hline
\end{tabular}
\captionof{table}{Scoring scheme to determine the traits based on text output.}
\label{tab:prob-table}
\end{table}

\subsection{Our Hypothesis for Personality Estimation}
Assessing these personality traits for humans typically involves using two types of data sources: self-reports and peer reports (e.g., friends, colleagues, etc.). Among the two, the more popular approach is via self‐reports, in which people describe how they see themselves while responding to a personality assessment questionnaire. For example, a participant is expected to respond to statements such as ``I am someone who is outgoing and sociable" on a Likert‐type scale (e.g., from $1$ = strongly disagree to $5$ = strongly agree). Self-reports capture people's explicit self-concepts about their personality traits, which are parts of their identities. On the other hand, peer reports help us understand how others perceive an individual. Unlike self-reports, peer reports focus on the personality traits attributed to an individual based on their interactions with others. To obtain peer reports, peers complete a similar personality assessment questionnaire about the individual, which then determines the perceived personality of that individual.

Our approach to measuring personality traits in language models will build on the human self-reporting approach, avoiding the design choices needed to qualify suitable peers. \emph{Our key hypothesis is that when prompted, language models generate text responses that reflect the personality traits of the datasets they were trained on.} Assuming that this hypothesis is true, we process the text responses generated by language models using auxiliary prediction models (which can be quite sophisticated themselves) to quantify their personalities.

\subsection{Zero-shot Classifier}
To the best of our knowledge, there is currently a dearth of textual datasets with personality-annotated data. Moreover, traditional machine learning approaches require vast amounts of data to achieve acceptable performance on prediction tasks. However, recent studies have shown promise in addressing the issue. For instance, the entailment-based ZSC method proposed by \citep{yin2019benchmarking} has demonstrated success in classifying tasks across multiple domains with satisfactory performance without requiring prior training data \citep{ma2021issues}. Therefore, we consider ZSC as our method for classifying personality traits. ZSC employs a pre-trained Natural Language Inference (NLI) model that utilizes both a \emph{premise} and a \emph{hypothesis} input. The model predicts whether the hypothesis is true (entailment), false (contradiction), or undetermined (neutral) with respect to the given premise.

ZSC uses the input text as the premise and generates a hypothesis for each potential label. The accuracy of each label is subsequently evaluated by determining if the NLI model predicts whether the premise "entails" the hypothesis. For example, let's consider the task of estimating the openness trait of a user based on their social media posts. We set the content of each post as the premise and construct a hypothesis for the \emph{openness} trait as: "This post reflects a high level of openness." We subsequently use the ZSC to evaluate the entailment and contradiction probabilities between the premise and the hypothesis. The resulting probabilities are then aggregated to obtain the likelihood of the user exhibiting high or low levels of the \emph{openness} trait. 

\subsection{Language Models in Open Ended Text Generation}\label{sec:lm}
We study multiple \emph{pretrained} language models that differ in their training strategies and corpora. These models rely on the concept of auto-regressive language generation, which assumes that the probability distribution of a word sequence can be decomposed into the product of conditional next-word distributions.
At the core of all these language models are neural network models known as Transformers ~\citep{vaswani2017attention}. Unlike traditional recurrent neural networks (RNNs), transformer architecture is based on a self-attention mechanism that allows the models to selectively focus on different parts of the input sequence while processing it. This framework is particularly beneficial for language understanding and text-generation tasks. Below, we briefly discuss the language models studied in the paper.

\bigskip
\noindent\textit{GPT-2}~\citep{radford2019language} is a transformer-based language model that is trained with a causal language modeling objective: predicting the next word given a sequence of previous words. GPT-2 was pretrained on the WebText dataset that was collected by scraping and filtering web pages from sources such as Reddit (a popular social networking website). 
\\~\\
\noindent\textit{GPT-3}~\citep{brown2020language} is the 3rd version release and is an upgraded version of GPT-2. It also uses a Transformer framework and Attention architecture similar to GPT-2. It is trained with $175$ billion parameters, which is over 10x the size of its predecessor, GPT-2. With its superior performance, it can generate text that human evaluators typically have a higher difficulty distinguishing from those written by humans. It was pretrained on an open-source dataset called \emph{Common Crawl}~\citep{wenzek2019ccnet} and other text corpora from sources such as Wikipedia (a popular online encyclopedia).
\\~\\
\noindent\textit{GPT-3.5-Turbo}~\citep{ye2023comprehensive} is a finetuned version of its predecessor, GPT-3. In addition to the datasets used in the training of GPT-3, developers leveraged human feedback to improve the language model's performance through a novel optimization framework known as reinforcement learning via human feedback. This unique approach to model training enables GPT-3.5-Turbo to provide more accurate and policy-optimized responses, making it a highly robust language model.
\\~\\
\noindent\textit{LLaMA (Language Model for Multi-task Architectures)} ~\citep{touvron2023llama} is a state-of-the-art collection of language models, with parameters ranging from 7B to 65B that are competitive with the best existing language models. These models are built by making modifications to transformer architecture proposed by ~\cite{vaswani2017attention} and trained using large amounts of textual data from various sources, including \emph{English CommonCrawl} (67\%), \emph{C4} (15\%)~\citep{raffel2020exploring}, \emph{Github} (4.5\%)~\citep{google-big-query}, \emph{Gutenberg} and \emph{Books3} (4.5\%)~\citep{gao2020pile}, \emph{ArXiv} (2.5\%)~\citep{lewkowycz2022solving}, and \emph{Stack Exchange} (2\%)~\citep{exchange2017stack}.
\\~\\
\noindent\textit{TransformerXL}~\citep{dai2019transformer} is a state-of-the-art language model developed by researchers at Carnegie Mellon University and Google AI Language. Its architecture is designed using new recurrence mechanisms and positional encoding schemes to overcome some of the limitations of the original Transformer architecture, such as the fixed-length context window and the difficulty in handling long-term dependencies. It was pretrained on the WikiText language modeling dataset, which was extracted from the set of verified \emph{good} and \emph{featured} articles on Wikipedia.
\\~\\
\noindent\textit{XLNet}~\citep{yang2019xlnet} is an extension of the TransformerXL model. The model learns bidirectional contexts by maximizing the expected likelihood over all permutations of the input sequence rather than just the preceding context. The auto-regressive objective provides a natural way to use the product rule for factorizing the joint probability of the predicted tokens, eliminating a specific independence assumption that was made in BERT~\citep{devlin2018bert}. This model was trained on BooksCorpus ~\citep{zhu2015aligning} and English Wikipedia datasets in a self-supervised fashion.

\section{Methods for Estimating and Altering Personalities}\label{sec:Methods}
In this section, we explore various approaches for evaluating the personality traits of datasets. Building on this, we design a robust novel approach to evaluate the traits of language models. Finally, we also present a few methods to alter the personality traits of language models.

\subsection{Evaluating Personalities of Datasets}

Extracting quantifiable personality traits from datasets requires defining suitable labels for ZSC, followed by a scoring scheme based on the ZSC outputs. As discussed earlier in Section \ref{sec:preliminaries}, ZSC takes a premise and a hypothesis as input and predicts whether the hypothesis is true (entailment), false (contradiction), or undetermined (neutral) given the premise. So the first step in using ZSC is to cast personality estimation as an entailment problem by converting the desired trait labels into hypotheses. In this work, we use the following hypothesis template while evaluating the ZSC: \emph{This response is characterized by $\{$label$\}$.} Here the placeholder $\{$\emph{label}$\}$ is replaced by trait-specific keywords as shown in Table~\ref{tab:extreme}.

We investigate different approaches to finalize our model setup and extract personality trait scores in terms of the Big Five factors. 

\paragraph{Approach 1}
The premise and the hypothesis can either be positively related or negatively related in this approach. Accordingly, we set up a single ZSC to quantify the five personality traits independently using five labels: \emph{agreeableness}, \emph{openness}, \emph{extraversion}, \emph{conscientiousness}, and \emph{emotional stability}. These labels will be used to generate hypotheses sequentially, and each will be passed as an input to the ZSC. We transform the resulting output probabilities (the entailment scores of the NLI model after normalization with contradiction scores) obtained for each label using one-dimensional linear interpolations to Big Five factor scores (ranging from $1$ to $5$).

\paragraph{Approach 2}
The premise and the hypothesis can either be positively related or unrelated under this approach. Under this assumption, we measure the personality scores using labels for the two extreme ends of each trait independently using five different ZSCs. Then, the output probabilities for both extremes are transformed via a softmax function to obtain the final scores for each Big Five factor. The labels for the extreme ends of each trait are provided in Table \ref{tab:extreme}.

\begin{table}
\scriptsize
\begin{minipage}{\linewidth}
\centering
\begin{tabular}{|l|c|}
\hline
  & \textbf{Labels}                    \\ \hline
1 & [agreeableness, antoganism]        \\ \hline
2 & [conscientiousness, disinhibition] \\ \hline
3 & [extraversion, introversion]       \\ \hline
4 & [emotional stability, neuroticism] \\ \hline
5 & [openness, closeness]              \\ \hline
\end{tabular}%
\captionof{table}{Labels for creating the input hypotheses for ZSC in Approaches 2 and 3.}
\label{tab:extreme}
\end{minipage} 
\end{table}

\paragraph{Approach 3}
In this approach, we consider all three scenarios, i.e., the premise and hypothesis can either be positively related, negatively related, or unrelated. Under assumptions made in the previous two approaches, measuring the five personality traits independently using ZSC might lead to incorrect results. The reason for this is that ZSC could classify a non-synonymic or antithetical hypothesis with low probability for a given premise. Therefore, one cannot guarantee if the lower score for a particular label is due to negative relation or no relation between the hypothesis and premise. Hence in this third approach, we set up a ZSC for each trait and measure the entailment for the two extreme ends in a dependent manner. Under the hood, the ZSC will calculate a single probability score using the underlying NLI's entailment scores of the two extreme ends of a trait. We then again match the probabilities to the big-5 scores. We use the same labels for the extreme ends of each trait as discussed in \textit{Approach 2}.

\paragraph{A Comparison Between the Three ZSC-based Approaches}
Our first approach works well whenever the hypothesis and premise entail or contradict each other. However, when the ZSC is prompted with unrelated premises and hypotheses, output label probabilities can be low, which leads to inaccurate measurements of personality traits. The second approach also has limitations. The ZSC uses the NLI model's entailment and contradiction scores behind the screens to calculate the probability of the output label. We cannot conclude that a low score implies that the hypothesis and the premise are neutral to each other, since the ZSC does not use the neutral score of the underlying NLI model. Our third approach overcomes these pitfalls of the first two approaches and is expected to provide a more precise assessment of the personality traits of datasets.

\subsection{Evaluating Personalities of Language Models} \label{ss:evaluate}

\begin{figure}[htbp]
\centering
\includegraphics[width=0.6\textwidth]{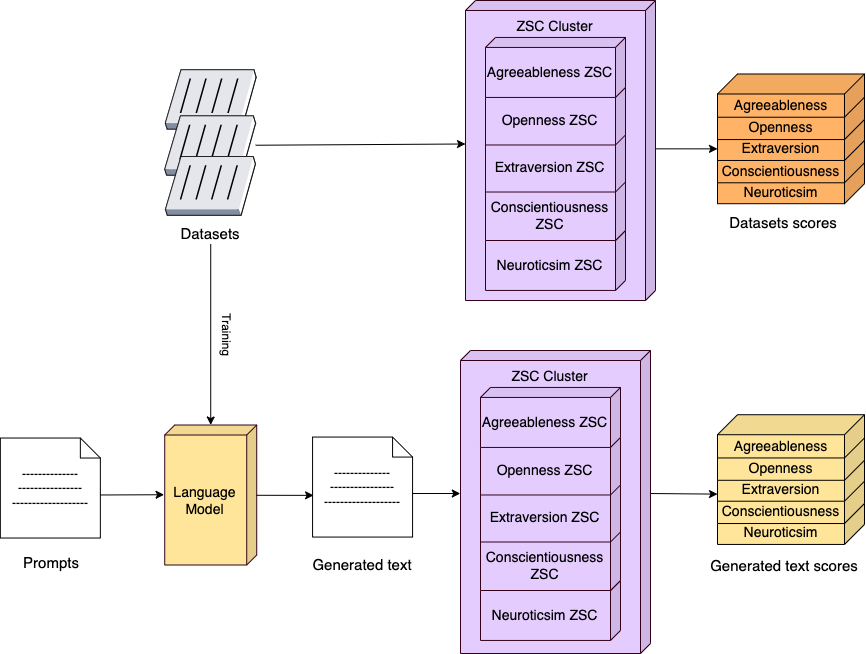}
\caption{\small{ Framework for evaluating the personality of training datasets and language models.}}
\label{fig:model}
\end{figure}

We adopt the assessment questionnaire for Big Five factors discussed in Section~\ref{sec:bff} as follows. 
Firstly, acquiring the responses in the multi-choice question-answer format is not directly feasible in open-ended text generation since the language model output is a sequence of words. Thus, we start by setting the statements from the questionnaire as prompts to the language model and generate text responses as shown in Figure~\ref{fig:model}. To account for the stochastic nature of the responses, we trigger the language model $N \in \mathbb{Z}_{+}$ times using the same prompt to observe $N$ different text completions for every statement in the questionnaire. Moreover,  we prompt the language models with each statement independently. This is to ensure that the order in which statements are fed as input to the model does not affect the personality assessment. 

Each of the generated text responses is passed independently as an input premise to the ZSC setup as discussed in \textit{Approach 3}. The resulting outputs are the independent probabilities for the responses to be characterized according to the five factors. We obtain $N$ such probabilities for each input prompt. We then use a one-dimensional linear interpolation to match the probabilities to scores ranging from 1 to 5. Finally, we compute the score distribution and summary statistics (such as the median) of the scores for all the statements to represent the personality of the concerned language model. 

\subsection{Modifying Personalities of Language Models} \label{modify_trait}
Pretrained language models possess varied personality traits due to their training on diverse datasets and due to differences in their model definitions and training approaches. We propose the following two model-agnostic methods aimed at altering the personality traits of language models.

\paragraph{\textbf{Method 1:}} \label{method1}  Modifying the personalities of the language models in a desired fashion can be viewed as updating the model's parameters such that it generates modified text responses that are closer to the desired personality traits. One way to partially achieve this is by finetuning the language models using suitably chosen personality-annotated text data. Although straightforward finetuning is an open-ended approach without allowing for precise control on achieving reference personality profiles, it allows the language model to partially adapt to the new data corpus in a straightforward manner and change the traits of the generated text without expending much computational overhead. Accordingly, when we trigger the finetuned language model with a prompt from our questionnaire, we expect that the generated response reflects the altered personality. In contrast, a precise control would need a new objective as well as a personality classifier in the loop, increasing the complexity of the process.

For finetuning, we leverage a (relatively smaller) personality-annotated text dataset made available as part of a machine learning competition \citep{siop:big5}. The dataset includes text responses to open-ended situational judgment items (SJIs) designed to elicit trait-relevant behaviors and aggregates trait scores based on the Big Five factors. To modify the personality of a language model with respect to a specific trait, we train the model on the filtered textual responses corresponding to that factor. Note that under this approach, precise control towards changing the personality traits to a specific desired set of values while maintaining language generation quality is generally non-trivial.
\begin{figure}[htbp]
\centering
\includegraphics[width=0.6\textwidth]{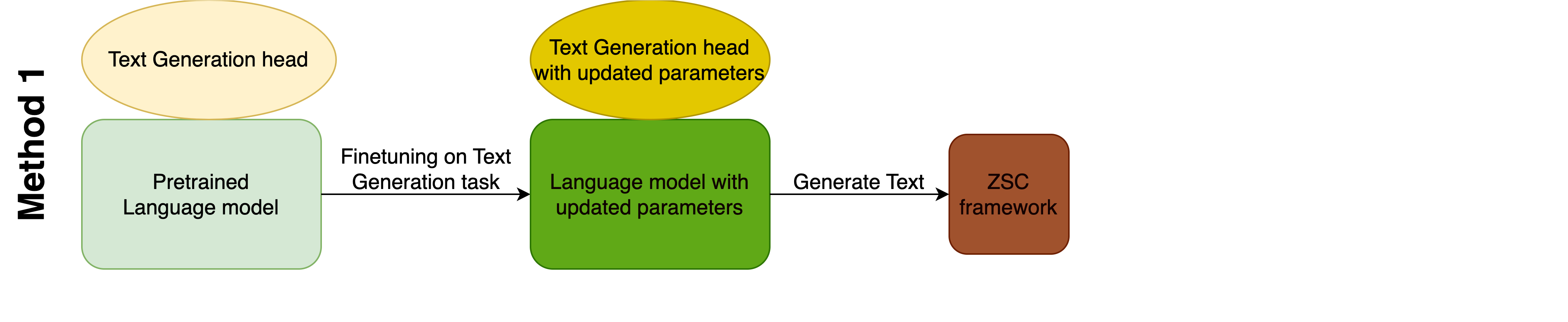}
\caption{Altering traits of language model using \textit{Method 1}.}
\label{fig:method1}
\end{figure}

\begin{figure}[htbp]
\centering
\includegraphics[width=0.6\textwidth]{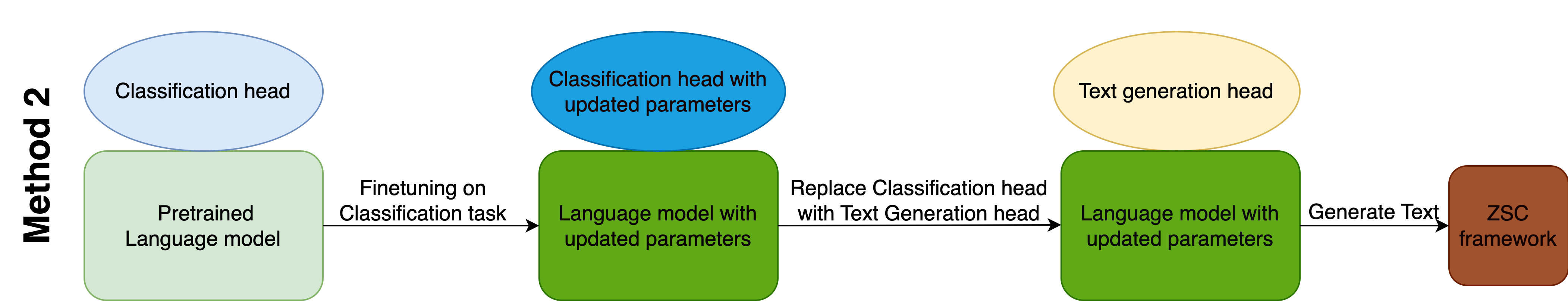}
\caption{Altering traits of language model using \textit{Method 2}.}
\label{fig:method2}
\end{figure}
\paragraph{\textbf{Method 2:}} \label{method2}  In this closely related approach to the above, we start by finetuning a pre-trained model on personality annotated text data but focus on a specific auxiliary classification task instead of the original text generation objective. In particular, using the text annotated with a specific Big Five factor,  we finetune the model on a binary classification task with opposite ends of a specific Big Five factor as target labels. Once we finetune the model via the auxiliary task, we use these updated model weights for text generation. 
Irrespective of the approach, we can evaluate the resultant model using the assessment process discussed in Section~\ref{ss:evaluate} and observe the altered traits.

\section{Experiments} \label{sec:experiments}
In this section, we present the experimental setup and the results obtained from evaluating the personality traits of the language models and the corresponding datasets they were trained on. In particular, we compare the personalities of language models and their underlying datasets to identify potential disparities. Finally, we report on the effectiveness of our approaches in altering the personality traits of language models.

\subsection{Traits of Datasets} \label{subsec:exp-datasets}

\paragraph{Setup}
We explore the personality traits of the following datasets used in training language models, namely:  \textsc{BookCorpus}, \textsc{English Wikipedia}, \textsc{WebText Test Set} and \textsc{Wikitext103}.

\begin{itemize}
\item \textsc{BookCorpus}~\citep{zhu2015aligning} is a large collection of free novel books written by unpublished authors, and contains $11,038$ books of $16$ different sub-genres and is used to train XLNet.

\item \textsc{English Wikipedia} contains cleaned articles that are built from Wikipedia content and used to train XLNet and GPT-3. However, the exact versions of the dataset used to develop those models are publicly undisclosed. We obtained a version of this data that was available on May $1^{st}, 2020$.

\item \textsc{WebText Test Set}~\citep{gokaslan2019openwebtext} is provided by the firm OpenAI. The training dataset was used to train GPT-2 and has not been publicly released. Hence, we use the test set for our experiments.

\item \textsc{Wikitext103}~\citep{merity2016pointer} contains more than $100$ million word tokens retrieved from the arrangement of (verified) good and featured articles on Wikipedia and is used to train TransformerXL.

\end{itemize}

Evaluating all of the datasets discussed above to infer personality traits requires extensive computational resources due to their large size. To overcome this challenge, we limit ourselves to analyzing sub-samples of the datasets, as specified in Table~\ref{tab:dataset-table}. These sub-sampled datasets can be evaluated at different levels of granularity, from sentence level and paragraph level to document level. Our analysis has revealed that when a document containing many paragraphs and sentences is passed as a standalone input, the ZSC that estimates personality traits doesn't perform well, predicting a score close to $3$ (which is neutral) for most samples. This is because such paragraphs and sentences may exhibit multiple conflicting traits or have multiple trait-less sentences (e.g., facts). To address this issue, we preprocess the data so that all samples are at the sentence or small paragraph level before evaluation. Once preprocessed, each sentence is passed as an input to ZSC using \textit{Approach 3} as explained in Section 2. Finally, we aggregate the resulting personality scores for each trait to obtain the personality trait distributions.

\begin{table}[]
\centering
\begin{tabular}{|c|c|c|c|c|}
 \hline
 Dataset & Size & \begin{tabular}[c]{@{}c@{}}Percent uses for\\inference\end{tabular} & Models\\
 \hline
 \textsc{Bookcorpus} & 5.75 GB  & 10\% & XLNet\\
 \hline
 \textsc{English Wikipedia} & 34.88 GB  & 2\% & GPT-3, XLNet\\
 \hline
 \textsc{Webtext Test Set} & 1.28 GB  & 20\% & GPT-2\\
\hline
 \textsc{Wikitext103} & 0.70 GB  & 100\% & Transformer-XL\\
 \hline
\end{tabular}
\caption{Summary statistics of the datasets.}
\label{tab:dataset-table}
\end{table}

\begin{figure}[htbp]
\centering
\begin{tabular}{cccc}
\includegraphics[width=.3\textwidth]{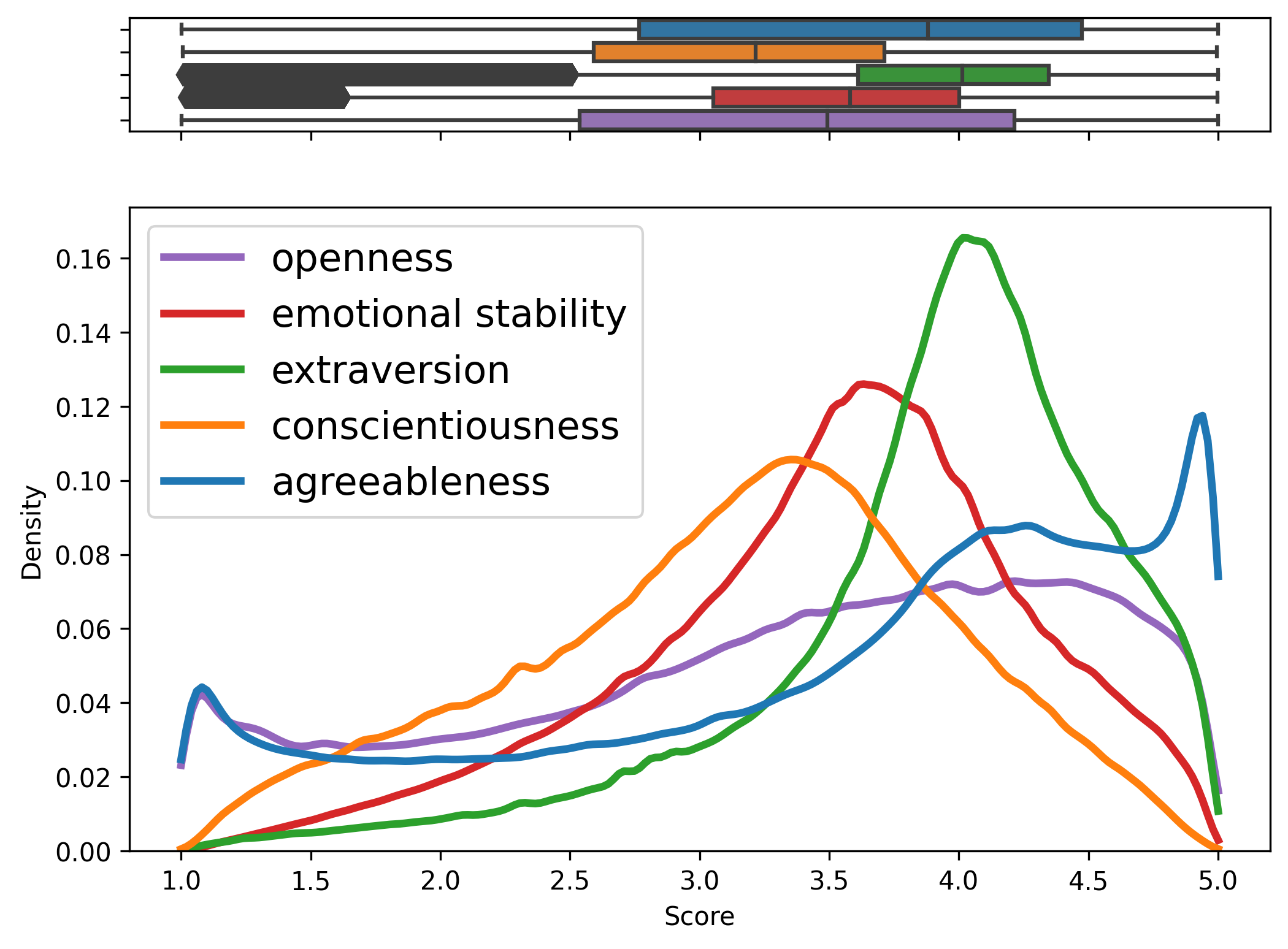}&
\includegraphics[width=.3\textwidth]{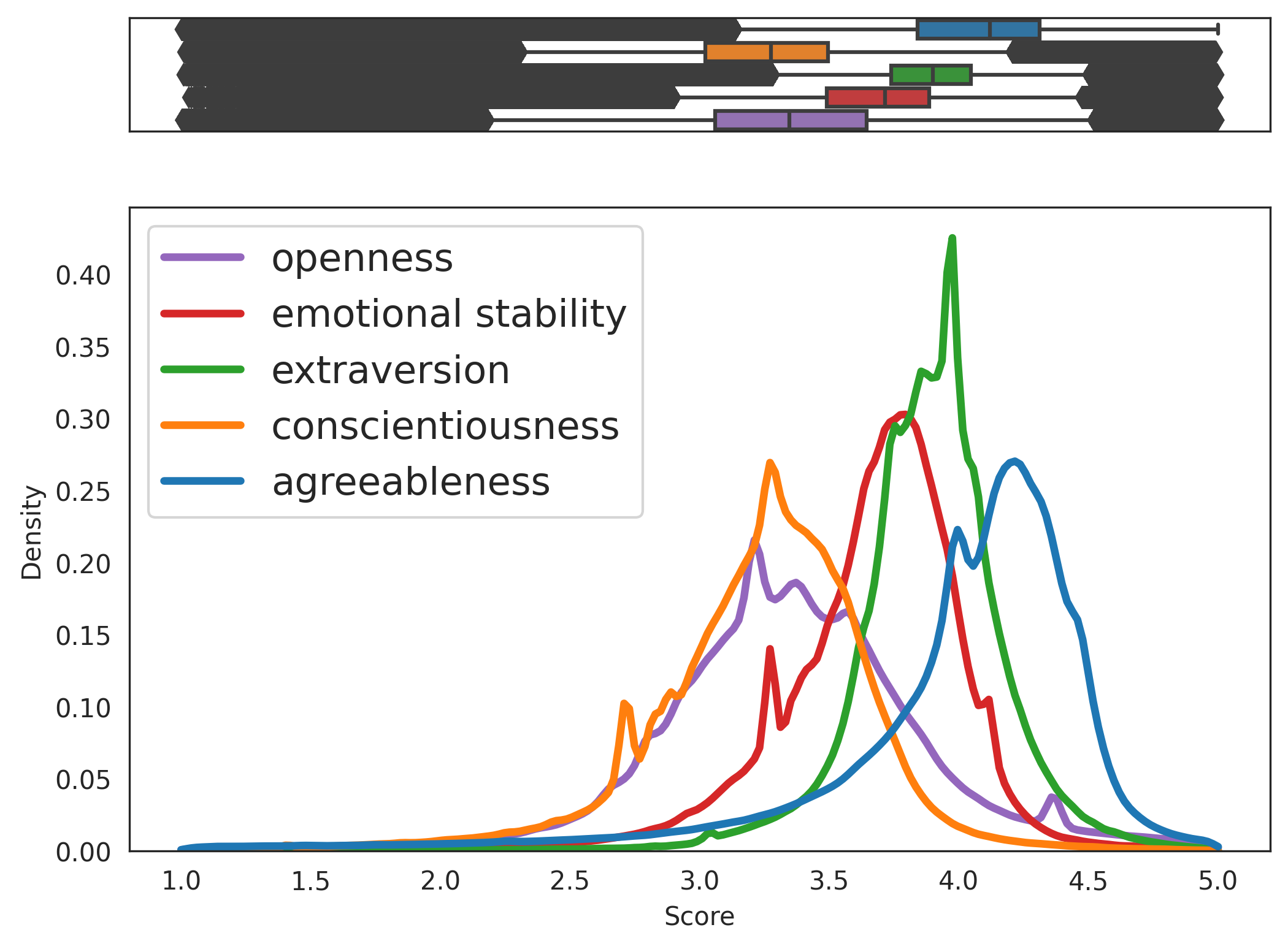}\\
(a) \small{\textsc{BookCorpus}}  & (b) \small{\textsc{English Wikipedia}} \\[6pt]
\end{tabular}
\begin{tabular}{cccc}
\includegraphics[width=.3\textwidth]{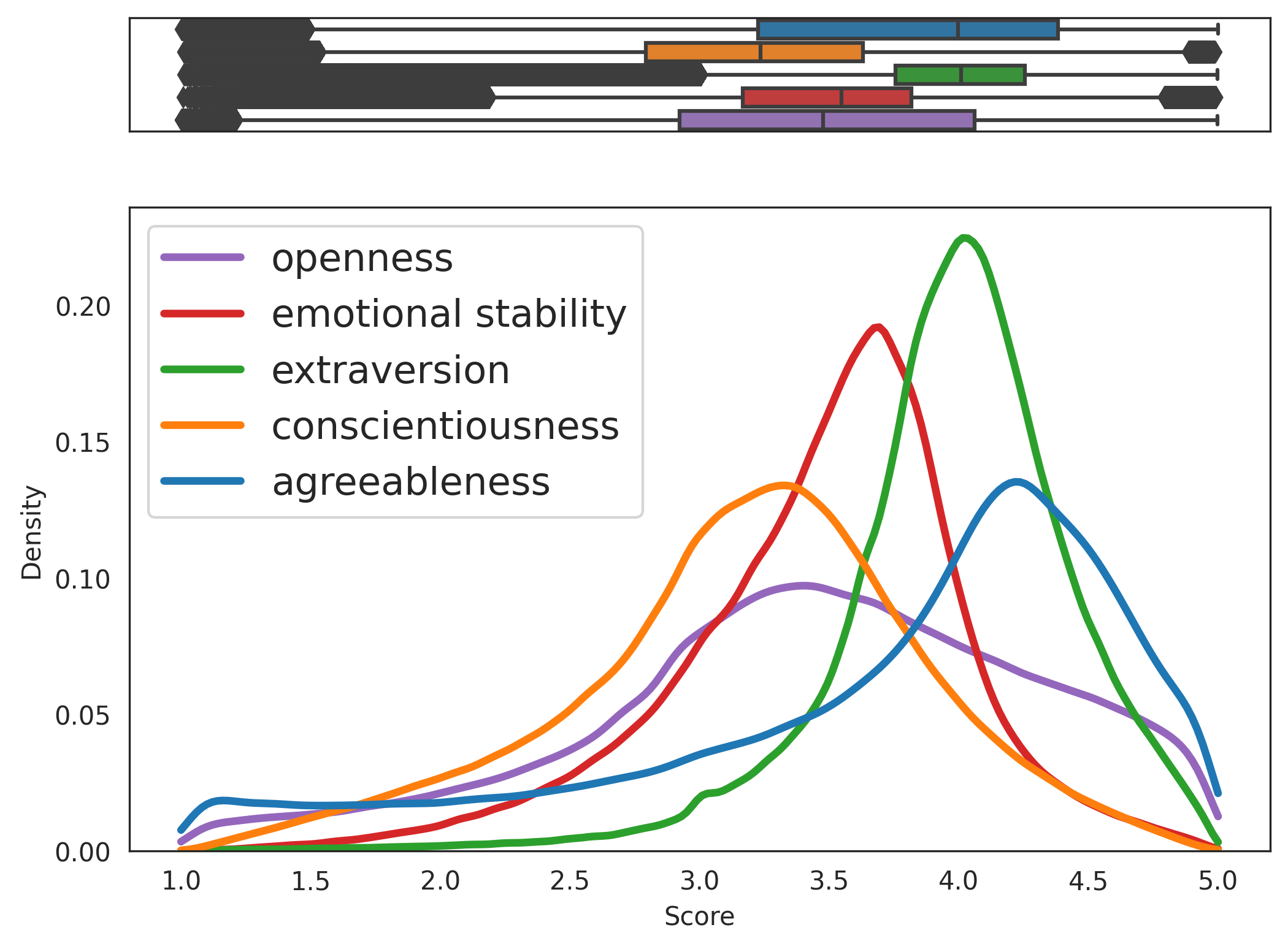} &
\includegraphics[width=.3\textwidth]{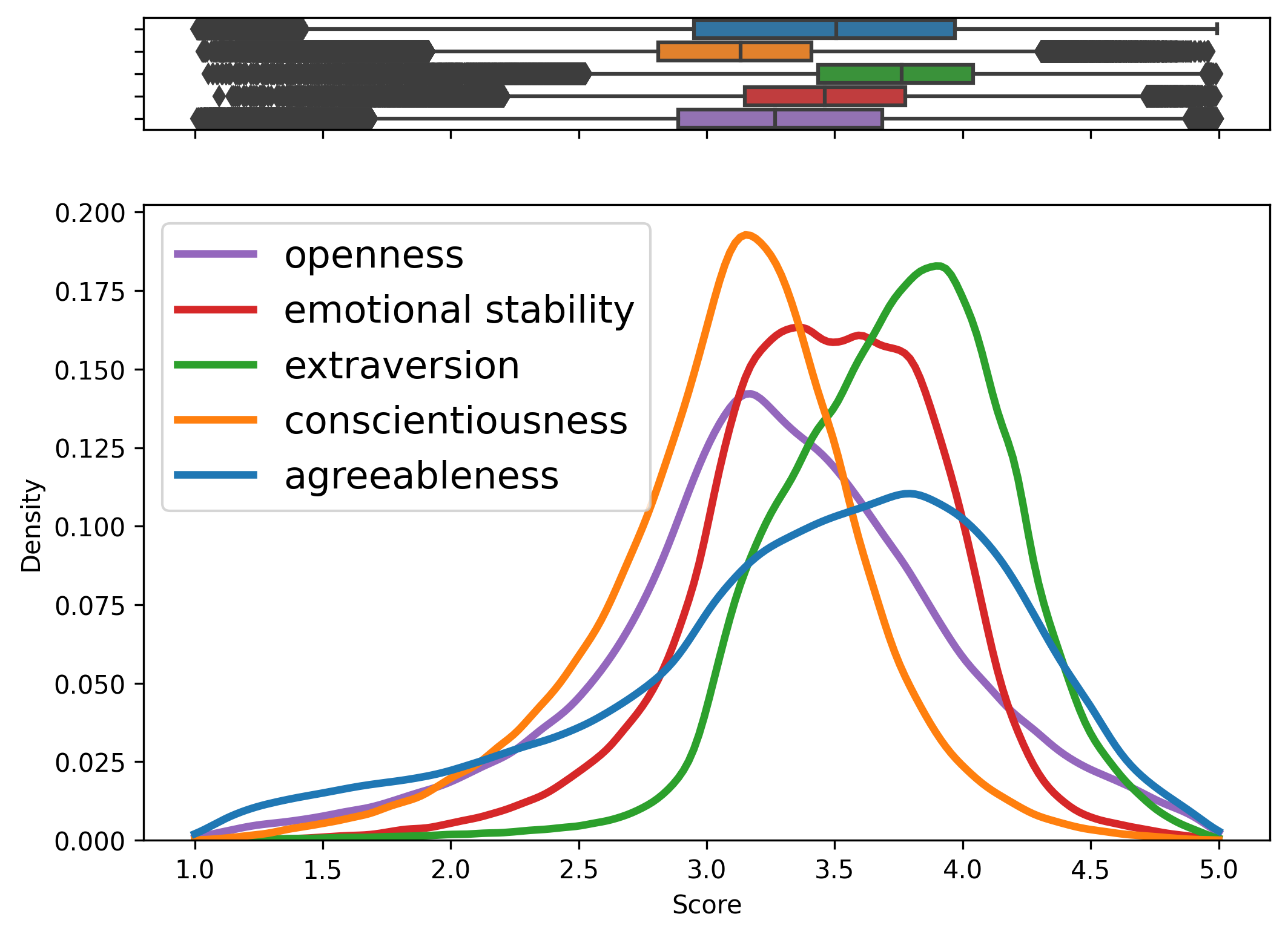} \\
(c) \small{\textsc{Webtext Test Set}}  & (d)\small{\textsc{Wikitext103}}  \\[6pt]
\end{tabular}
\caption{Personality trait distributions of datasets.}
\label{fig:dist_datasets}
\end{figure}

\paragraph{Results}
Figure~\ref{fig:dist_datasets} presents the estimated distributions of personality traits across the datasets referenced earlier. Overall, we observe trait distributions are positively skewed, as reflected by scores greater than the neutral $3.0$. Additionally, to aid the comparison of these distributions, we report the median scores in the adjoining box plots, in which the length of the boxes represent the variance of the traits.

We make a few key observations next. In the \textsc{BookCorpus} dataset, \emph{Extraversion} and \emph{Agreeableness} are the most prominent traits, followed by \emph{Openness}, \emph{Emotional Stability}, and \emph{Conscientiousness}. Further, \emph{Agreeableness} scores have the highest variance, followed by \emph{Openness}. For the \textsc{English Wikipedia} dataset, the most prominent trait is \emph{Agreeableness}, followed by \emph{Extraversion}, \emph{Emotional Stability}, \emph{Openness}, and \emph{Conscientiousness}. This dataset has smaller spreads compared to others. In the \textsc{WebText Test Set} dataset, \emph{Agreeableness} and \emph{Extraversion} are again the prominent traits, followed by \emph{Emotional Stability}, \emph{Openness}, and \emph{Conscientiousness}. The spreads of the distributions of \emph{Agreeableness} and \emph{Openness} are wider than the other traits in this dataset. \emph{Extraversion} is the most prominent trait in the \textsc{Wikitext103} dataset, while \emph{Conscientiousness} is the least prominent. \emph{Agreeableness} scores have the widest spread in this dataset, followed by \emph{openness}. The personality traits exhibited by the datasets discussed above align with the expected characteristics from their content. For instance, the \textsc{Wikitext103} dataset contains a wealth of factual information, resulting in predominantly neutral personality traits. By contrast, the skewed distribution of traits observed in the \textsc{BookCorpus} dataset reflects its inclusion of diverse text genres, such as fiction, non-fiction, and academic material. The trait profiles could have been acquired through alternative techniques; however, we employ ZSCs to maintain consistency with the estimation method used for the models. 

\subsection{Traits of Models} 
\label{subsec:traits_models}
\paragraph{Setup}
To quantify the personality traits of language models, we trigger them using the Big Five questionnaire $30$ times. This is done to account for the stochastic nature of the responses and to obtain a reliable and consistent estimate of personality traits. The generated text responses are then classified using the ZSC framework discussed in \textit{Approach 3}.

We have used the publicly available OpenAI programmatic API To evaluate the performance of GPT-3 and GPT-3.5-Turbo. Among the LLaMA models, we evaluated its publicly available 4bit-7b quantized version. All three models discussed above operate as black box models due to their undisclosed architecture. Consequently, our evaluation focuses solely on comparing their personality traits without the ability to modify them. We utilized the text generation pipeline from Hugging Face's \texttt{transformer} package ~\citep{wolf2019huggingface} to evaluate these white-box language models, namely GPT-2, TransformerXL, and XLNet. To ensure the consistency of our experiments, we employed a standardized set of hyper-parameters for text generation for evaluating each of the language models, i.e., we chose \textit{maxseqlength} (maximum length of output sequence) as $256$, \textit{temperature} (randomness parameter) as $1$, and a \textit{top-p} (smallest possible set of words whose cumulative probability exceeds the probability p) as $1$.

\begin{table}[]
\centering
\resizebox{\textwidth}{!}{%
\begin{tabular}{|l|l|}
\hline
\textbf{Prompt} &
  \textbf{Generated Response} \\ \hline
I don’t talk a lot. &
  \begin{tabular}[c]{@{}l@{}}I'm more of an introvert so I'm less than happy to take time to talk. \\ I prefer to be able to take control over how I work and make sure I be able to talk a lot.\end{tabular} \\ \hline
I’m relaxed most of the time. &
  \begin{tabular}[c]{@{}l@{}}In all my experiences, I did not notice a single problem about myself or others I dealt with, \\ because I was focused on my personal well-being and that was all …\end{tabular} \\ \hline
I feel comfortable around people. &
  \begin{tabular}[c]{@{}l@{}}I get a lot done and it's nice working out or just enjoying myself and being around people. \\ A lot of that comes with me …\end{tabular} \\ \hline
I am full of ideas. &
  \begin{tabular}[c]{@{}l@{}}I am also a very creative person, so it is important to me to develop my own ideas. \\ I have a lot of inspiration and a lot of love for the world.\end{tabular} \\ \hline
\end{tabular}%
}
\vspace{3mm}
\caption{Examples of (uncorrected) text responses generated using GPT-2.}
\label{tab:gpt2-text}
\end{table}

\paragraph{Results}
Table \ref{tab:gpt2-text} shows a few examples of input prompts and the respective text responses generated using GPT-2. Due to the auto-regressive nature of the text generation process, we observe that the GPT-2 model generates contextually coherent text based on the input prompts. 

\begin{figure}[htbp]
\captionsetup[subfigure]{font=scriptsize,labelfont=scriptsize}
\centering
\begin{multicols}{3}
    \subcaptionbox{GPT-2}{\includegraphics[width=\linewidth]{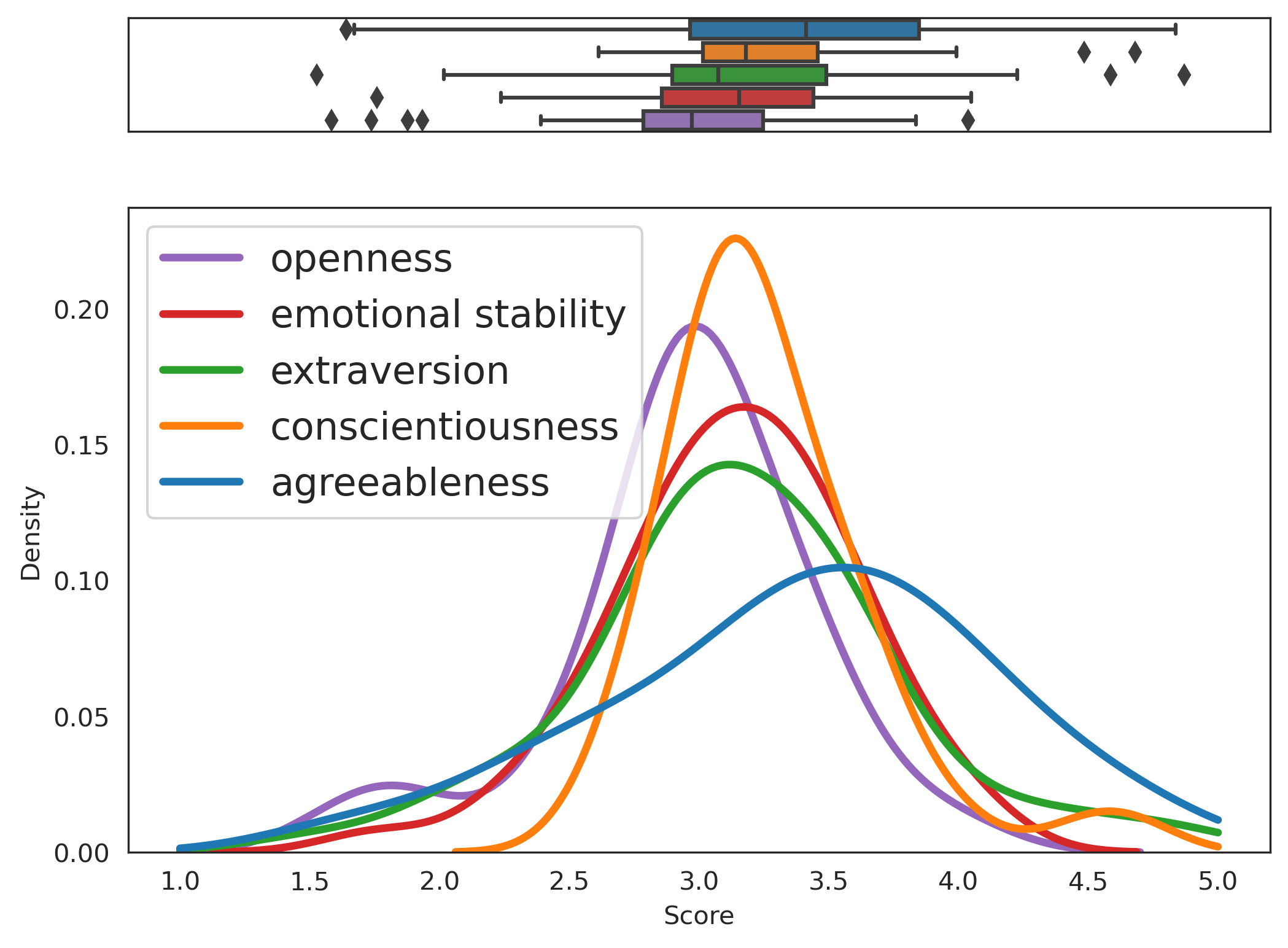}}    \par
    \subcaptionbox{GPT-3}{\includegraphics[width=\linewidth]{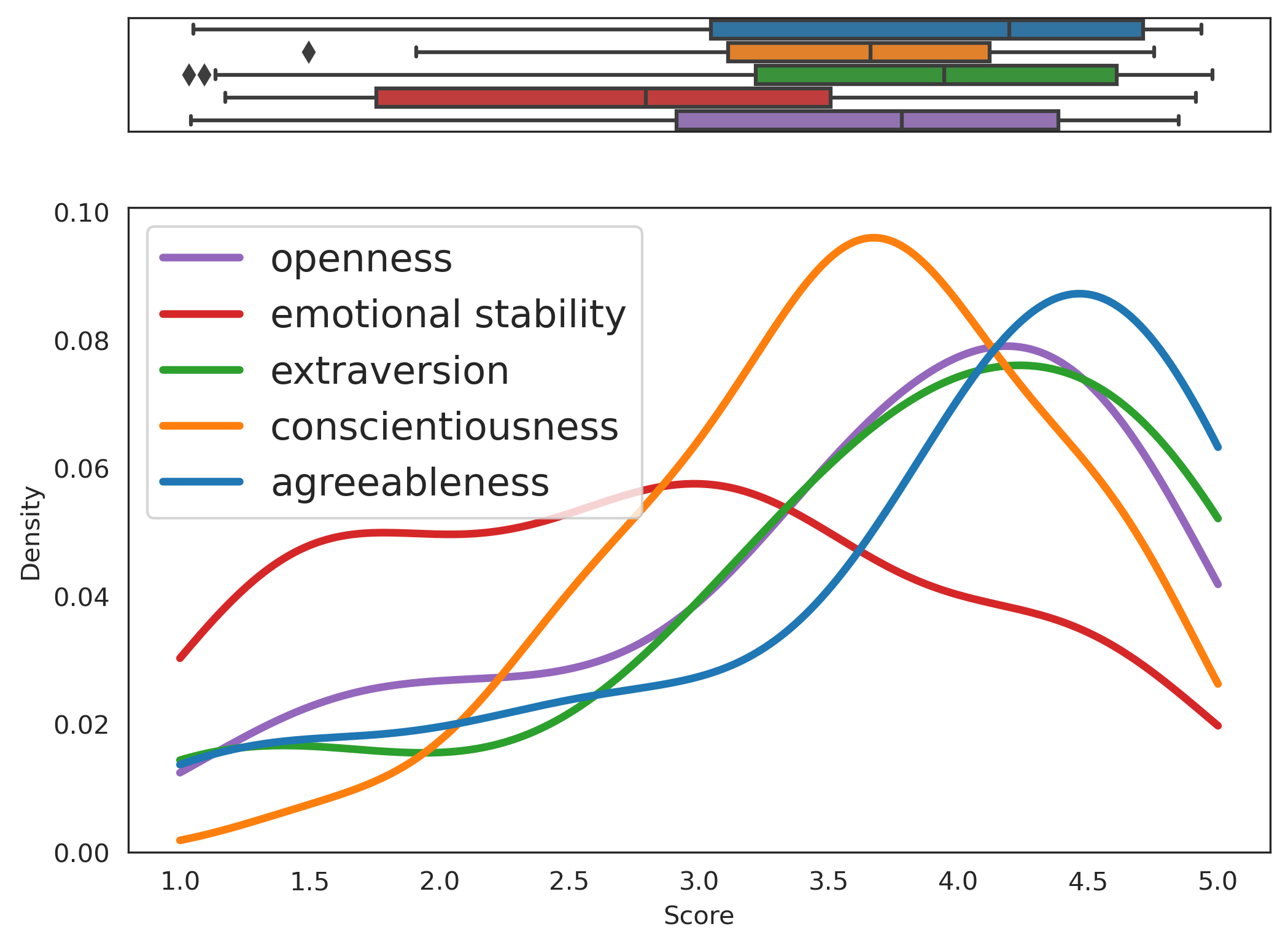}}\par
    \subcaptionbox{GPT-3.5-Turbo}{\includegraphics[width=\linewidth]{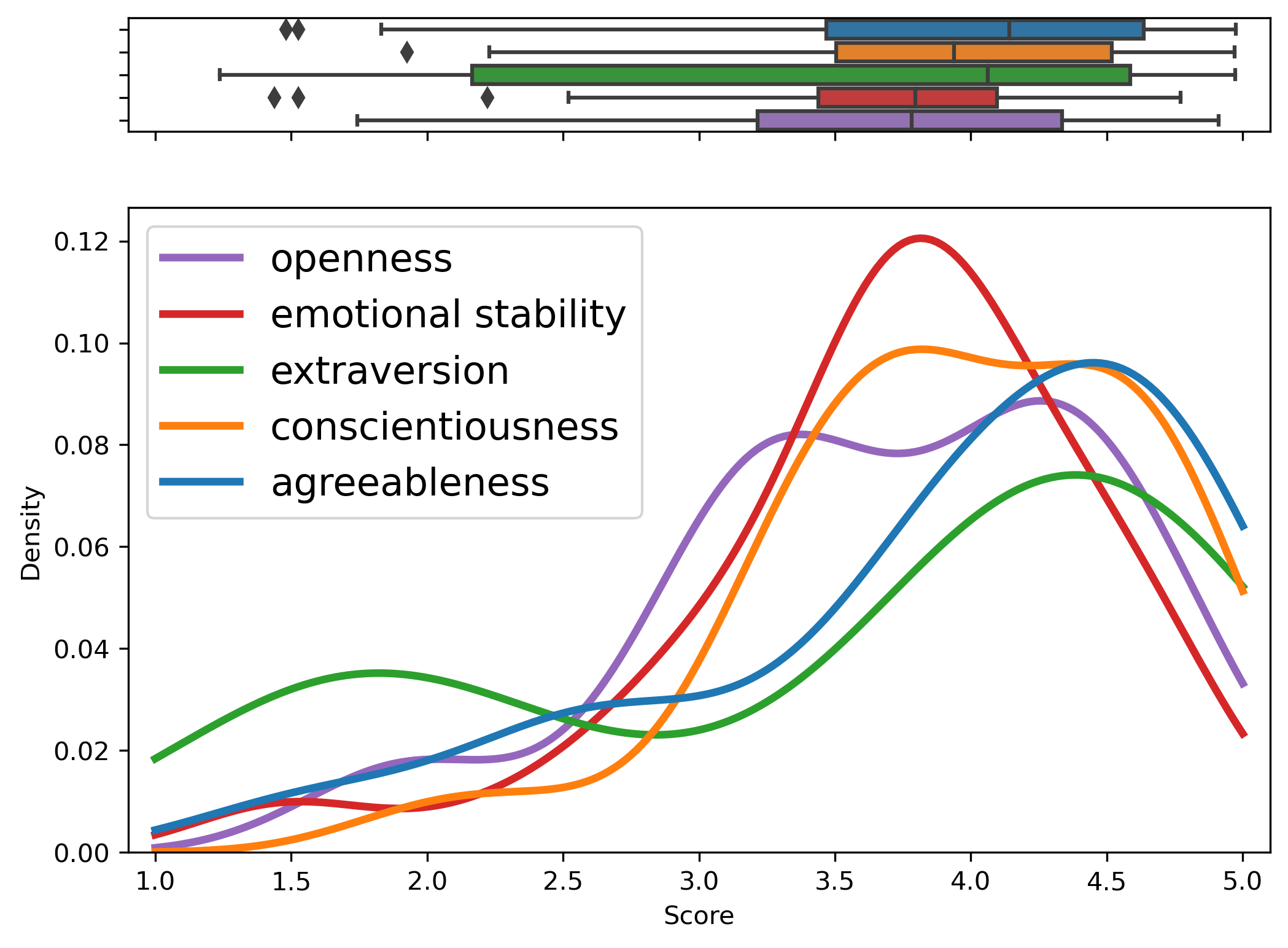}}\par
\end{multicols}
\begin{multicols}{3}
    \subcaptionbox{TransformerXL}{\includegraphics[width=\linewidth]{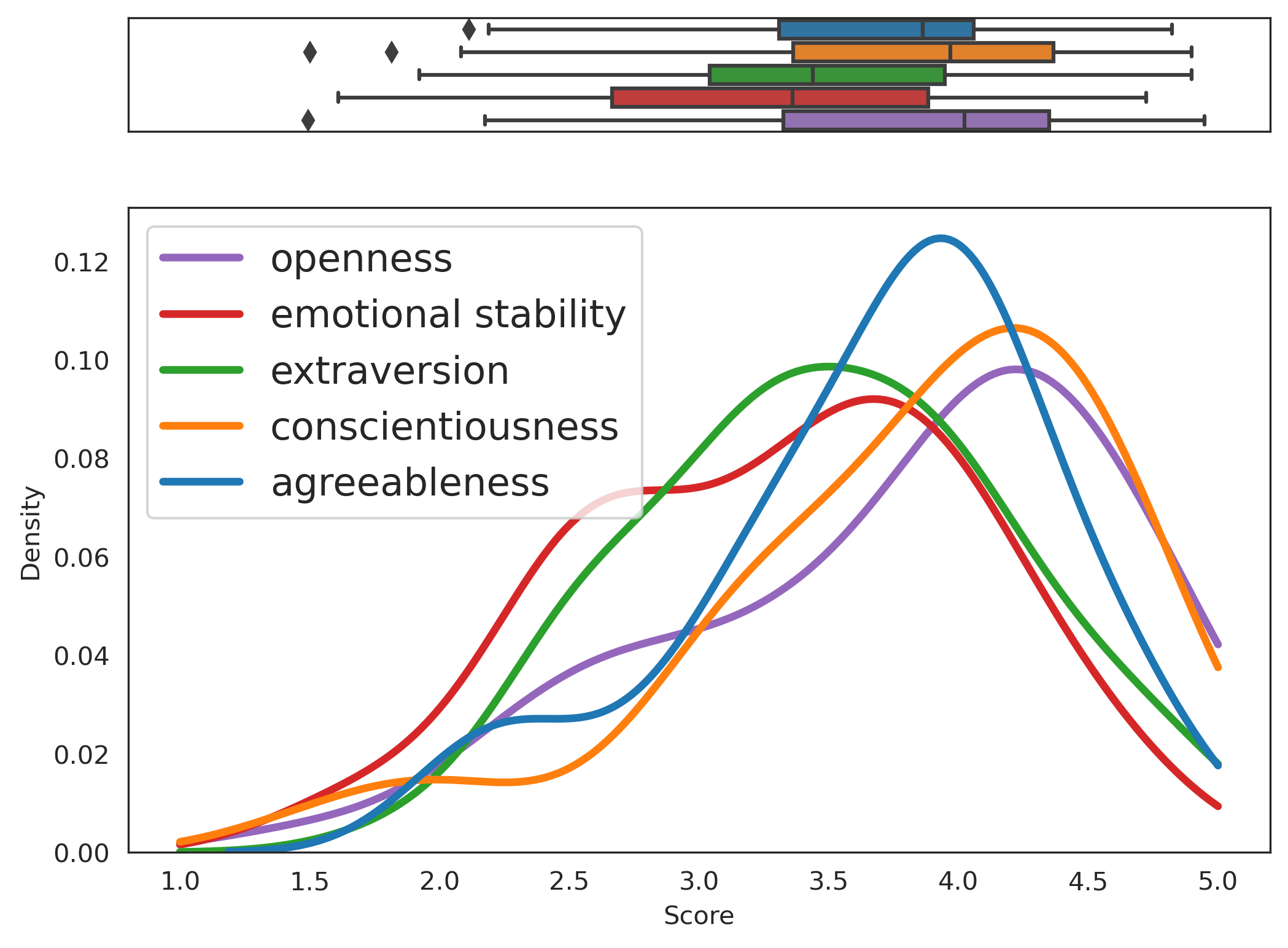}}\par
    \subcaptionbox{XLNet}{\includegraphics[width=\linewidth]{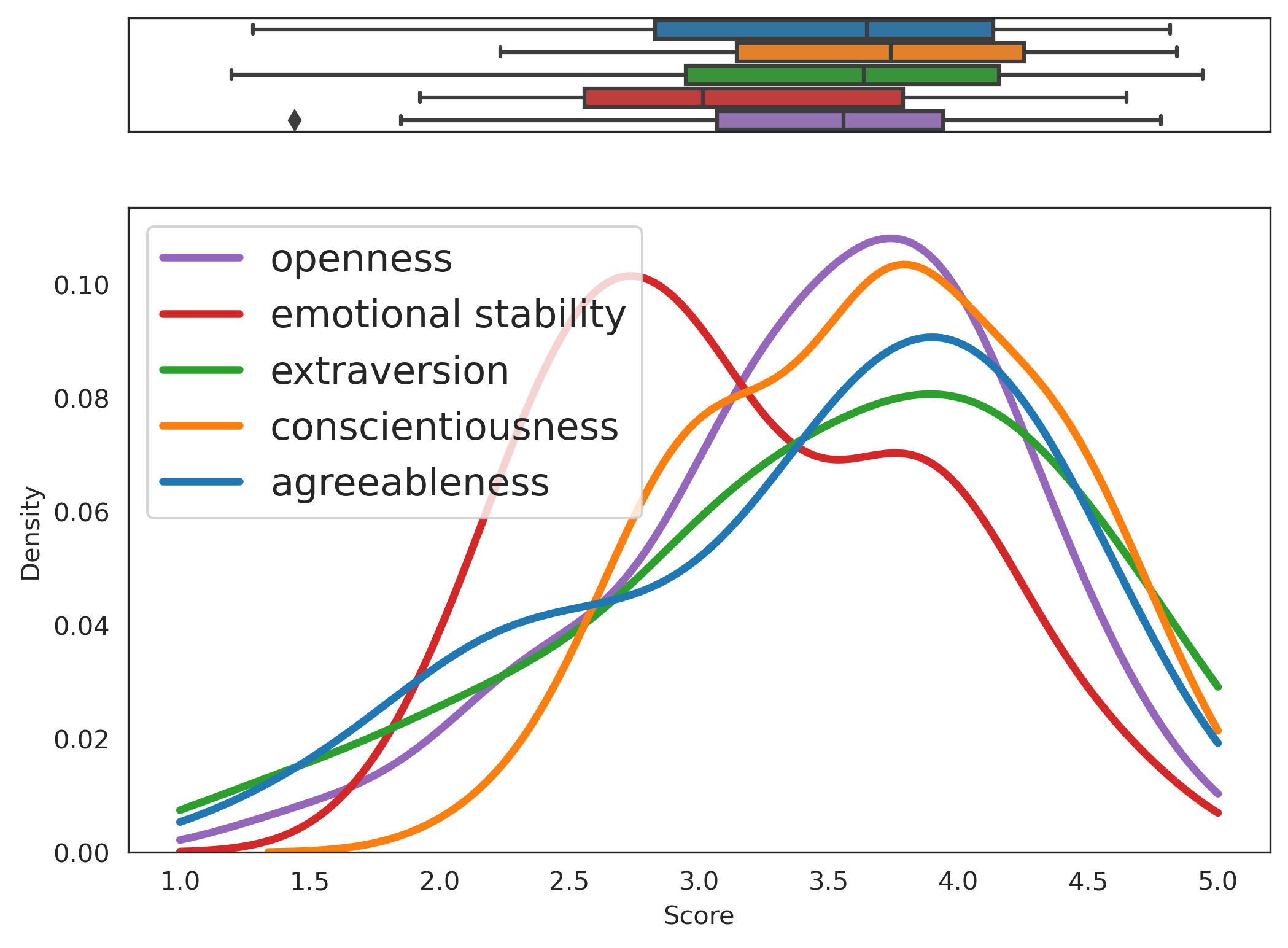}}\par
    \subcaptionbox{LLaMA}{\includegraphics[width=\linewidth]{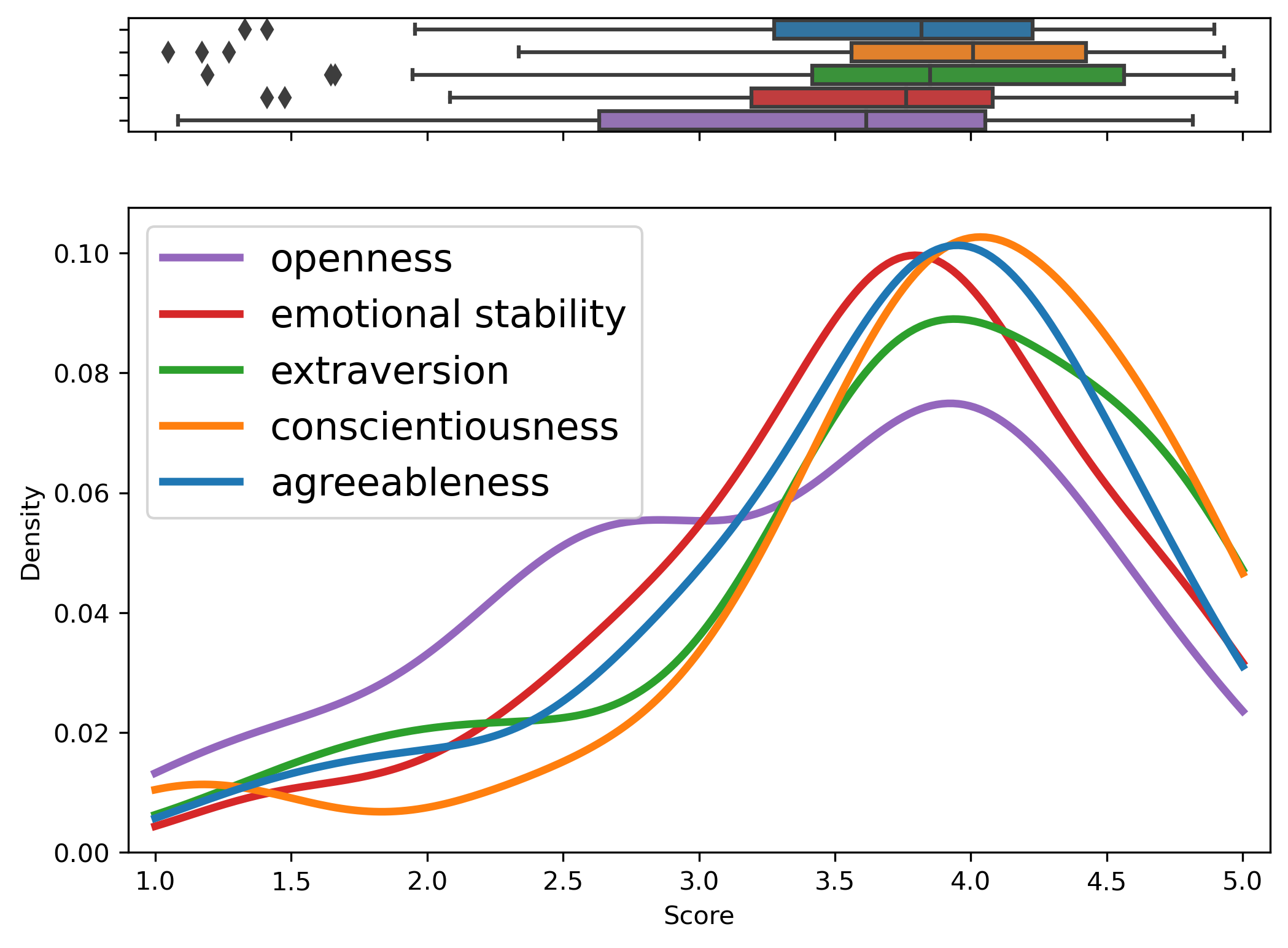}}\par
\end{multicols}
\caption{Personality trait distributions of language models.}
\label{fig:trait_eval_dist}
\end{figure}

\begin{figure}[htbp]
\centering
\begin{tabular}{cccc}
\includegraphics[width=.3\textwidth]{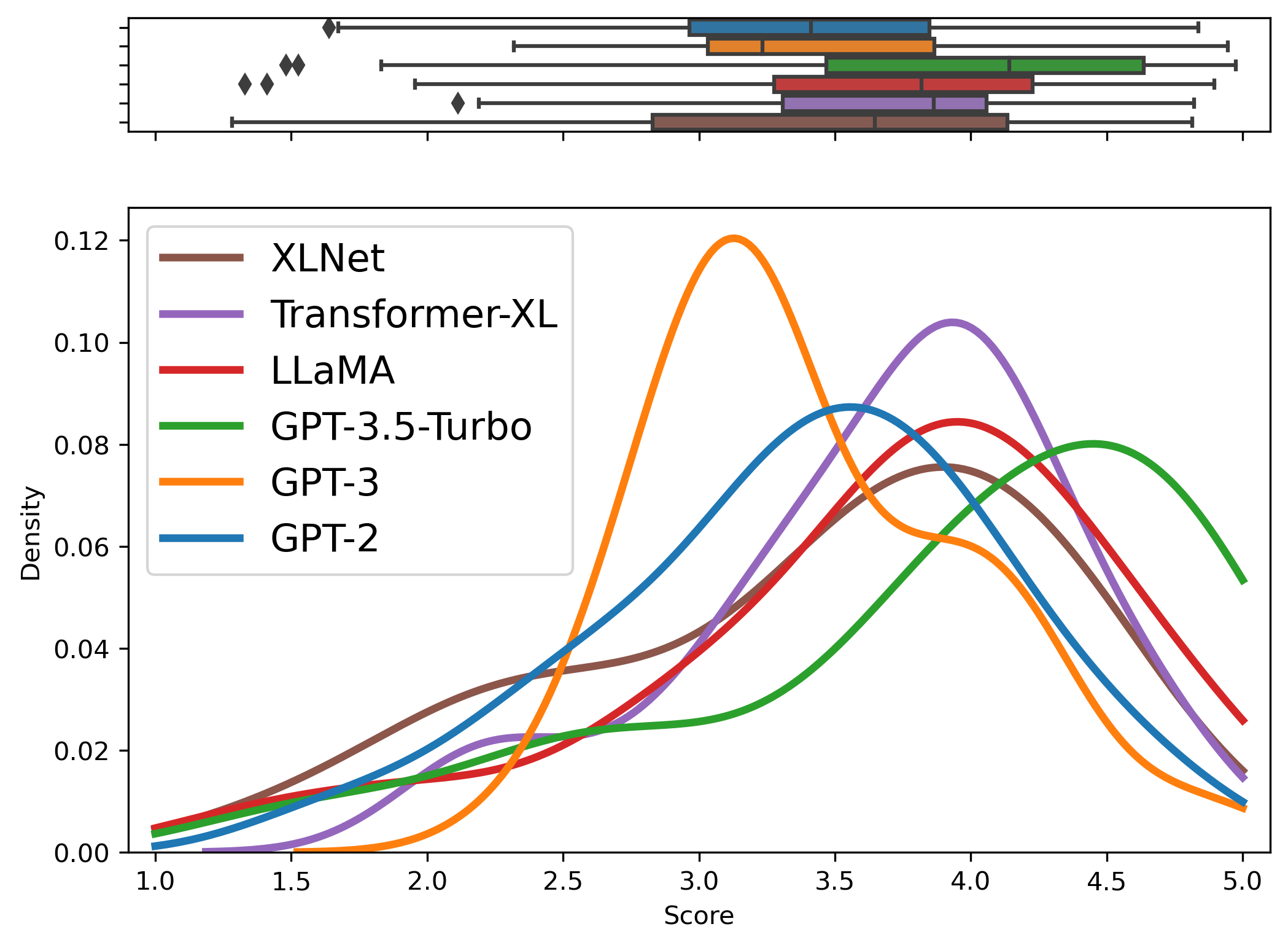}&
\includegraphics[width=.3\textwidth]{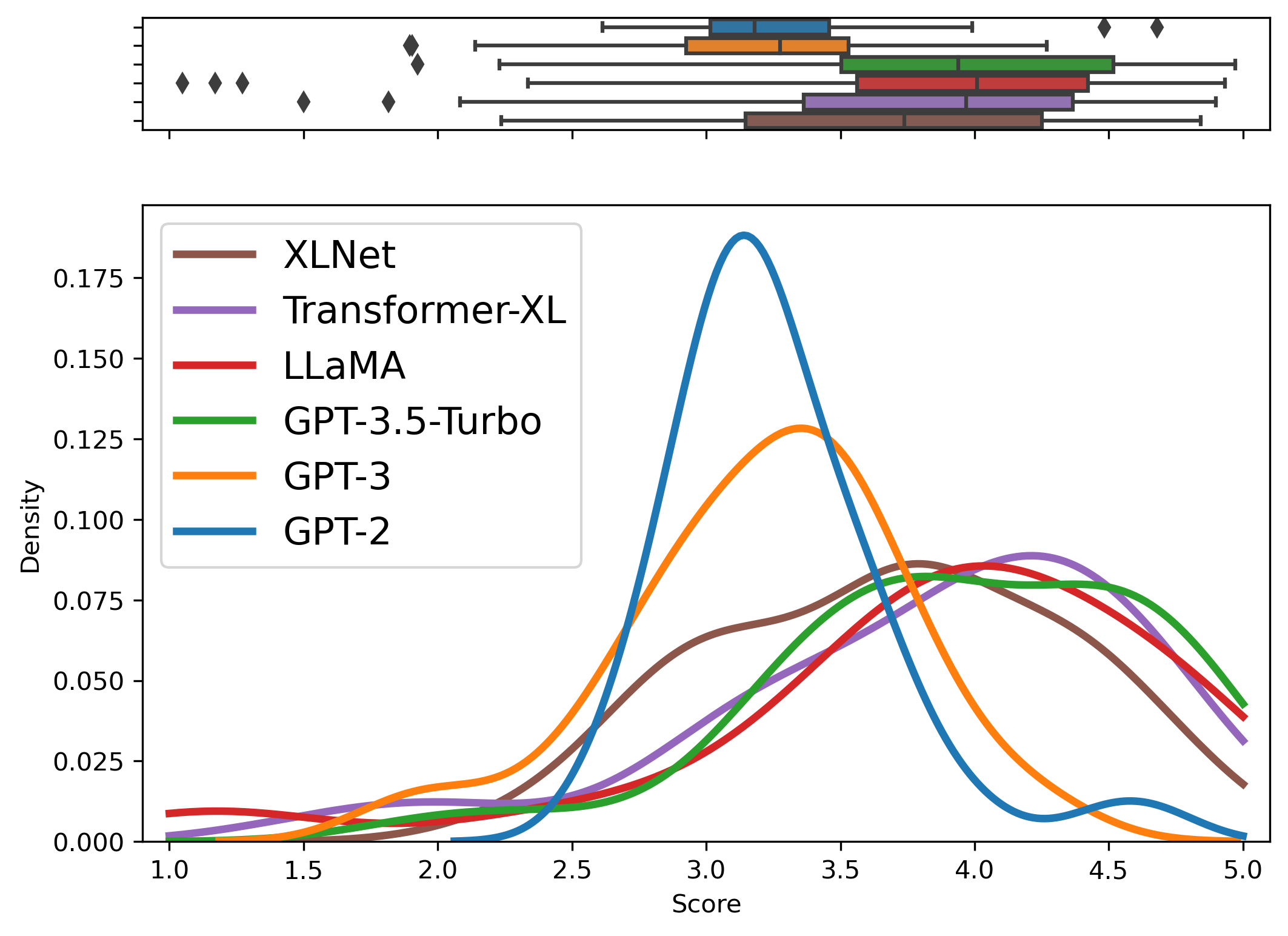}&
\includegraphics[width=.3\textwidth]{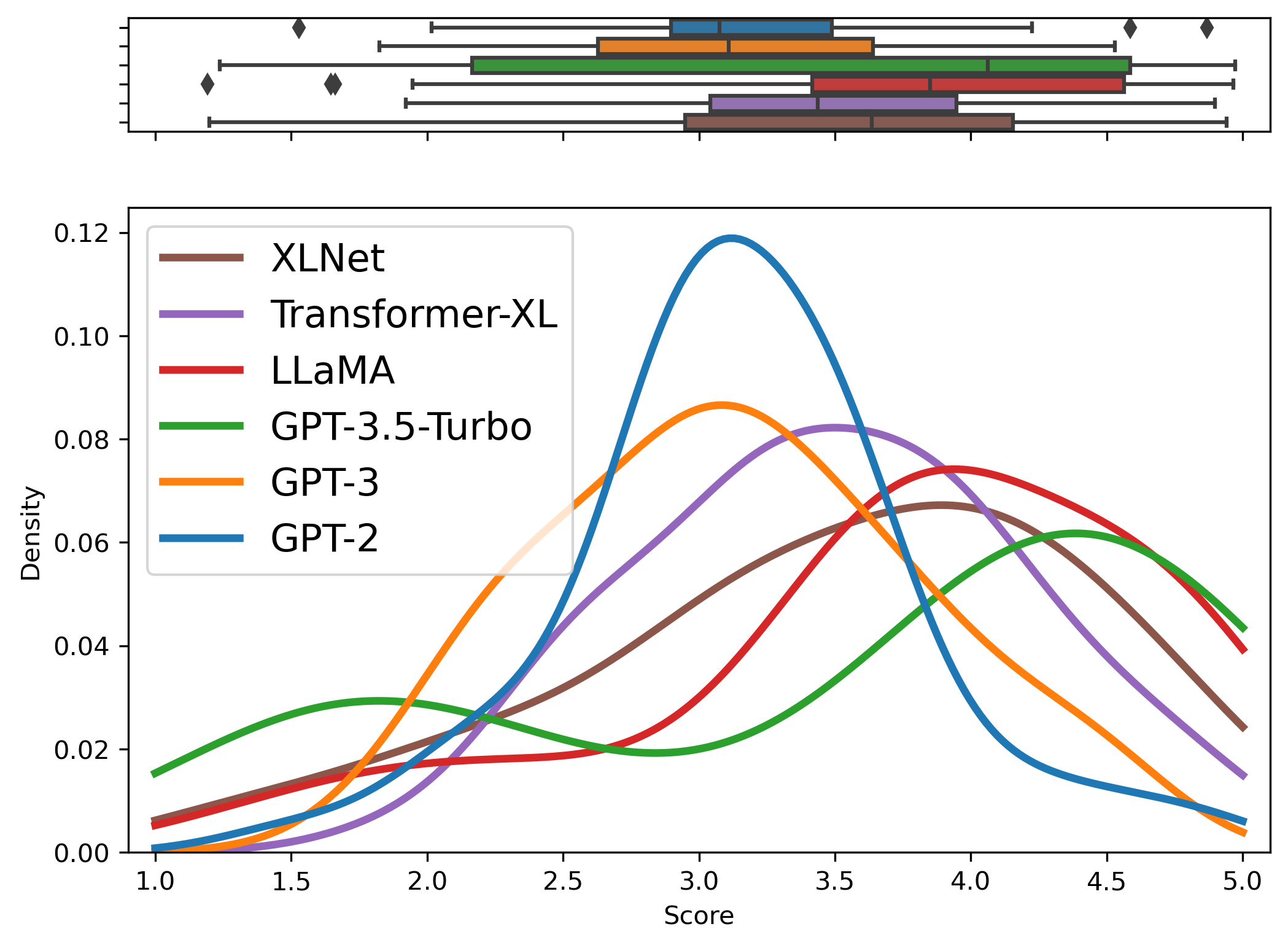} \\
(a) \small{Agreeableness}  & (b) \small{Conscientiousness} & (c) \small{Extraversion}  \\[6pt]
\end{tabular}
\begin{tabular}{cccc}
\includegraphics[width=.3\textwidth]{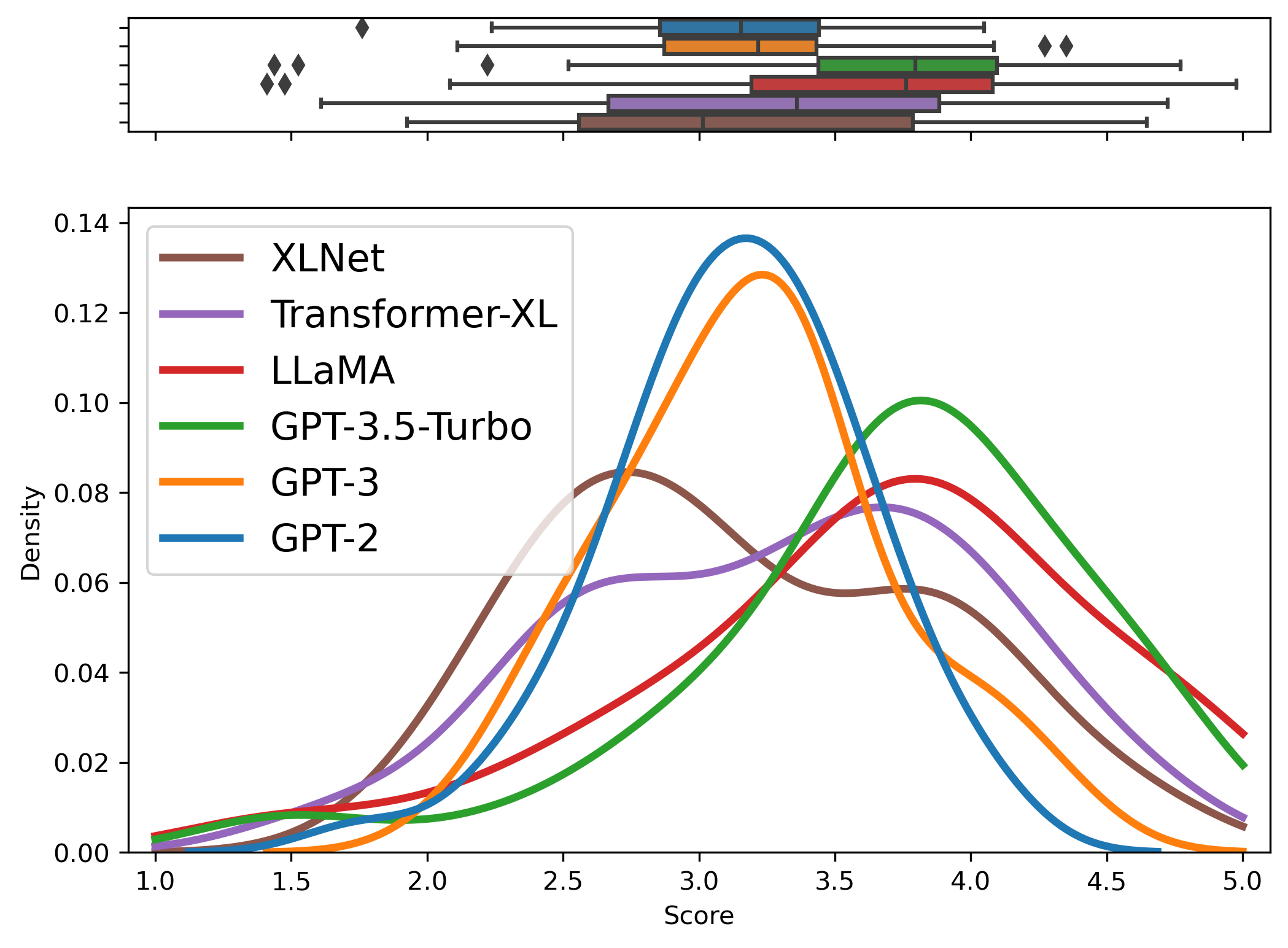} &
\includegraphics[width=.3\textwidth]{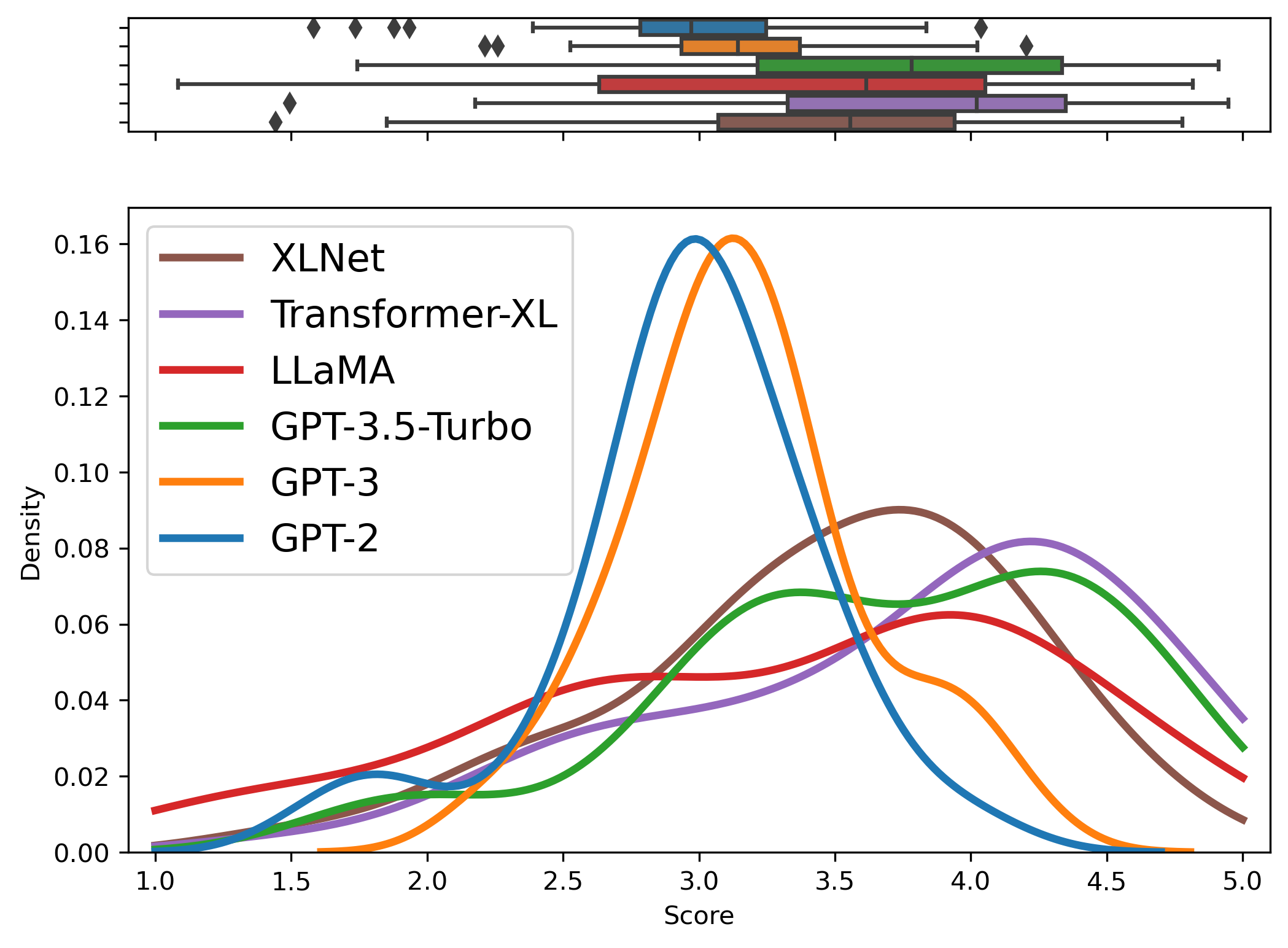} \\
(d) \small{Emotional Stability}  & (e) \small{Openness}  \\[6pt]
\end{tabular}
\caption{Comparison of personality trait distributions across language models.}
\label{fig:model_compare}
\end{figure}

We observe noticeable variations in trait distributions of text generated using different language models, as illustrated in Figures \ref{fig:trait_eval_dist} and \ref{fig:model_compare}. These variations are also evident from the median scores of personality traits, as shown in Table \ref{tab:eval-table}. Based on these scores, we can make a few key observations. Firstly, the \emph{Agreeableness} scores are higher for GPT-3.5-Turbo, suggesting that the generated text for this language model reflects a more generous and empathetic personality. Secondly, the LLaMA model scores higher in \emph{Conscientiousness}. On the other hand, GPT-3.5-Turbo has a higher median \emph{Extraversion} score, implying that the generated responses emulate extroverted rather than introverted personalities. Finally, the LLaMA model scores the highest in \emph{Emotional Stability}, reflecting a more emotionally stable personality than the other language models under consideration. All the language models have their unique set of inherent personality traits, which can be leveraged in downstream applications based on specific use cases. For instance, the LLaMA model, with its high emotional stability median score, is suitable for designing a chatbot in a mental health care setting, facilitating more empathetic conversations with users. Overall, Our work's significant contribution is the principled approach to extracting and utilizing these personality traits to enhance the effectiveness of language models in practical applications.
\\

\begin{table}[htbp]
\centering
\resizebox{\textwidth}{!}{%
\begin{tabular}{|c|c|c|c|c|c|c|}
\hline
Trait             & GPT2        & GPT3        &GPT-3.5-Turbo &LLaMA  &Transformer-XL &XLNet       \\ \hline
Agreeableness     & 3.41 (0.73) & 4.19 (1.14) & 4.41 (0.93) & 3.82 (0.85) & 3.86 (0.66)    & 3.64 (0.87) \\ \hline
Conscientiousness & 3.18 (0.39) & 3.66 (0.77) & 3.94 (0.69) & 4.01 (0.90) &3.96 (0.78)    & 3.73 (0.64) \\ \hline
Extraversion      & 3.07 (0.60) & 3.94 (1.10) & 4.06 (1.24) & 3.85 (0.94) &3.43 (0.69)    & 3.63 (0.91) \\ \hline
Emotional Stability       & 3.15 (0.46) & 2.79 (1.11) & 3.79 (0.73) & 3.76 (0.82)&3.36 (0.74)    & 3.01 (0.70) \\ \hline
Openness          & 2.97 (0.47) & 3.78 (1.06) & 3.78 (0.79) & 3.62 (1.00) &4.02 (0.83)    & 3.55 (0.71) \\ \hline
\end{tabular}%
}
\vspace{3mm}
\caption{Personality scores (along with their uncertainties) of Language Models.}
\label{tab:eval-table}
\end{table}

\noindent\textit{Sensitivity of the estimated personality distributions to estimation parameters:}
\\
As discussed earlier, the datasets used for training these language models are composed of text at different levels of granularity. Similarly, the text responses generated using these models can be multiple sentences long. This variation in text length can affect the personality trait distributions that we observe. To assess the sensitivity of estimation to this, we generate personality trait scores using different output modes, as explained below.
\begin{itemize}
    \item \textit{Mode 1:} Trait score of the entire generated response.
    \item \textit{Mode 2:} Trait score of the first sentence in the generated response.
    \item \textit{Mode 3:} Median of the trait scores of all sentences present in the generated response.
\end{itemize}

\begin{figure}[htbp]
\captionsetup[subfigure]{font=scriptsize,labelfont=scriptsize}
\begin{multicols}{3}
    \subcaptionbox{GPT-2 (\textit{Mode 1})}{\includegraphics[width=\linewidth]{gpt2.png}}\par
    \subcaptionbox{GPT-2 (\textit{Mode 2})}{\includegraphics[width=\linewidth]{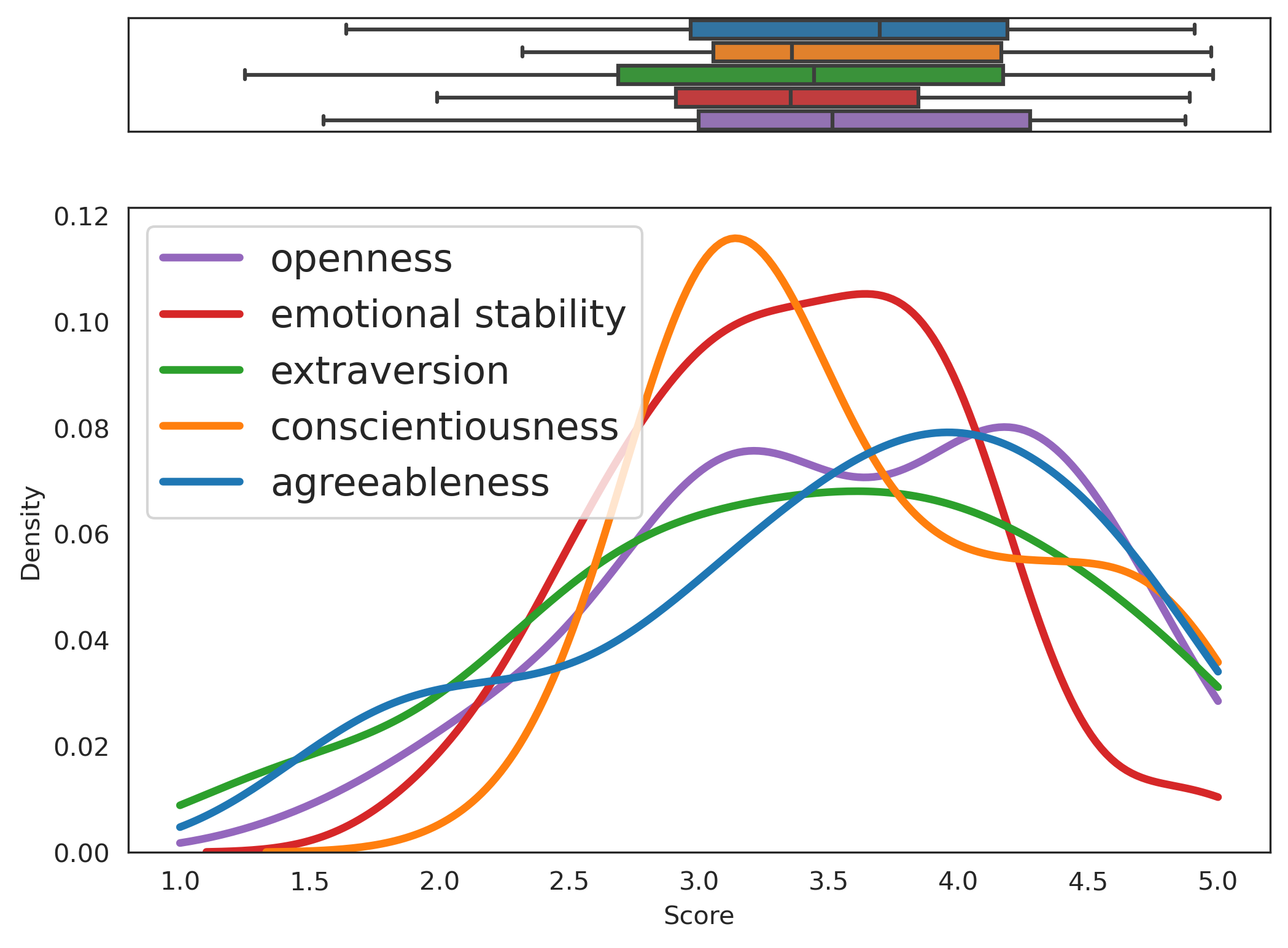}}\par
    \subcaptionbox{GPT-2 (\textit{Mode 3})}{\includegraphics[width=\linewidth]{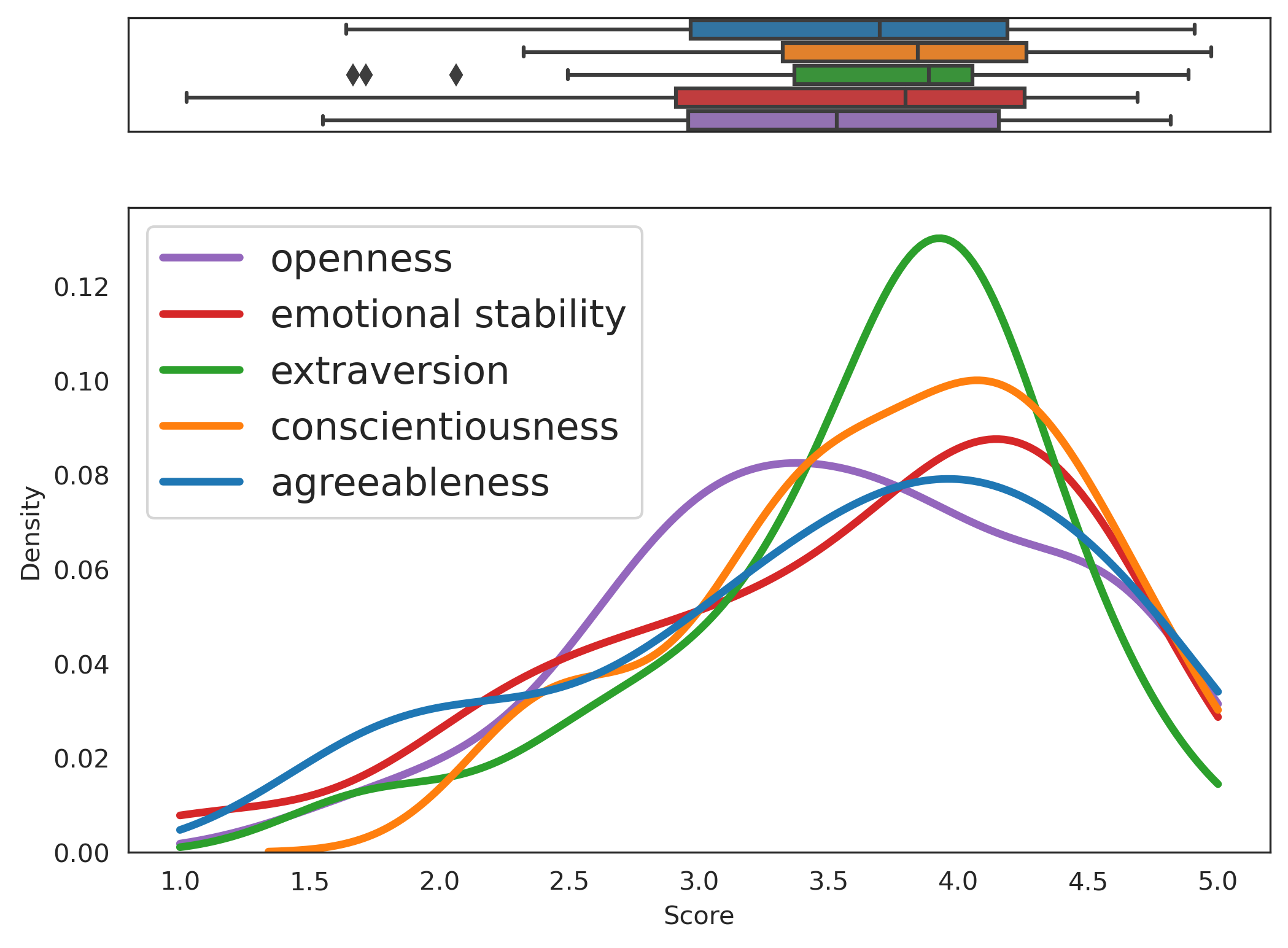}}\par
\end{multicols}
\begin{multicols}{3}
    \subcaptionbox{GPT-3 (\textit{Mode 1})}{\includegraphics[width=\linewidth]{gpt3.png}}\par
    \subcaptionbox{GPT-3 (\textit{Mode 2})}{\includegraphics[width=\linewidth]{gpt3.png}}\par
    \subcaptionbox{GPT-3 (\textit{Mode 3})}{\includegraphics[width=\linewidth]{gpt3.png}}\par
\end{multicols}
\begin{multicols}{3}
    \subcaptionbox{TransformerXL (\textit{Mode 1})}{\includegraphics[width=\linewidth]{transfoxl.png}}\par
    \subcaptionbox{TransformerXL (\textit{Mode 2})}{\includegraphics[width=\linewidth]{transfoxl.png}}\par
    \subcaptionbox{TransformerXL (\textit{Mode 3})}{\includegraphics[width=\linewidth]{transfoxl.png}}\par
\end{multicols}
\begin{multicols}{3}
    \subcaptionbox{XLNet (\textit{Mode 1})}{\includegraphics[width=\linewidth]{xlnet.png}}\par
    \subcaptionbox{XLNet (\textit{Mode 2})}{\includegraphics[width=\linewidth]{xlnet.png}}\par
    \subcaptionbox{XLNet (\textit{Mode 3})}{\includegraphics[width=\linewidth]{xlnet.png}}\par
\end{multicols}
\caption{Personality trait distributions of language models obtained from different modes of evaluation.}
\label{eval_models}
\end{figure}

Figure \ref{eval_models} shows the trait distributions for some of the language models obtained using the modes listed above. The distributions remain the same for GPT-3, XLNet, and TransformerXL. This is due to the fact that the responses generated using these models were single sentences. However, the trait distributions vary for GPT-2 across different modes of evaluation, implying that the structure of the output influences the personality score estimation. This observation can be made quantitative using Wasserstein distances (between trait distributions across models) and is provided in Table~\ref{tab:wd_gpt_modes}. From the table, we observe that \emph{Agreeableness} varies the least, i.e., is least sensitive to how much of the response is used to assess the trait. And while there is a lot of change across all these traits, \emph{Openness} varies the most.

\begin{table}[]
\scriptsize
\centering
\begin{tabular}{|c|c|c|}
\hline
\textbf{Trait} &
  \textbf{\begin{tabular}[c]{@{}c@{}}Wasserstein distance \\ Mode 2 vs Mode 1\end{tabular}} &
  \textbf{\begin{tabular}[c]{@{}c@{}}Wasserstein distance \\ Mode  3 vs Mode 1\end{tabular}} \\ \hline
Agreeableness       & 0.248 & 0.248 \\ \hline
Conscientiousness   & 0.563 & 0.361 \\ \hline
Extraversion        & 0.489 & 0.425 \\ \hline
Emotional Stability & 0.559 & 0.249 \\ \hline
Openness            & 0.591 & 0.610 \\ \hline
\end{tabular}
\caption{Comparing traits distributions across modes using Wasserstein distances $\in [0,1]$ for GPT-2.}
\label{tab:wd_gpt_modes}
\end{table}

\begin{figure}[htbp]
\captionsetup[subfigure]{font=scriptsize,labelfont=scriptsize}
\begin{multicols}{3}
    \subcaptionbox{\textsc{WebText Test Set} }{\includegraphics[width=\linewidth]{webtext_inf}}\par
    \subcaptionbox{GPT-2 evaluated considering input prompts from Big Five Questionnaire.}{\includegraphics[width=\linewidth]{gpt2_256_split_first.png}}\par
    \subcaptionbox{GPT-2 evaluated without considering input prompts.}{\includegraphics[width=\linewidth]{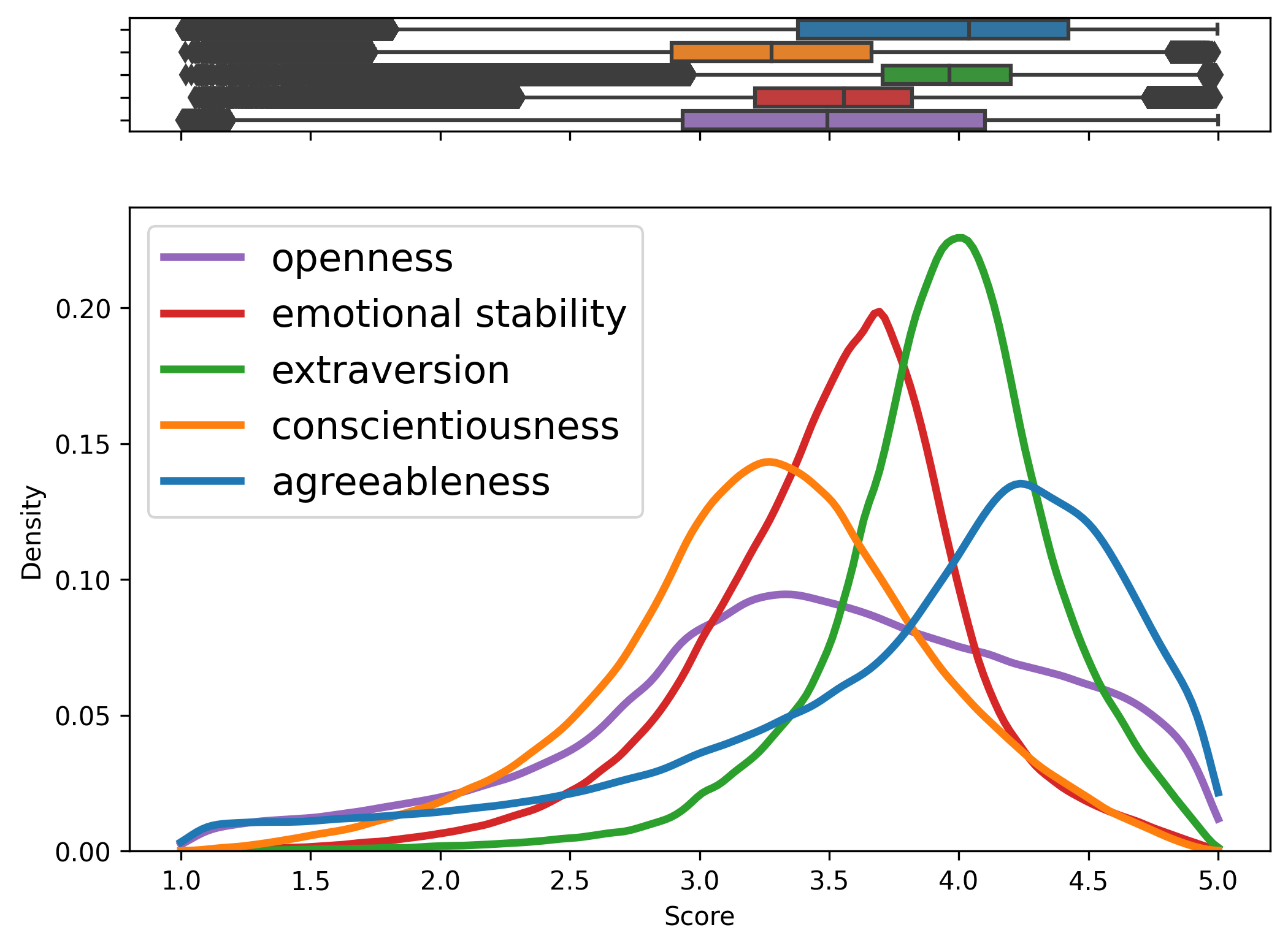}}\par
\end{multicols}
\caption{Distributions of different traits in: (a) \textsc{WebText Test Set} , and (b) GPT-2.}
\label{fig:dist_comp}
\end{figure}

\subsection{Traits of Datasets vs Models}
To validate our hypothesis of whether the language models inherit personality traits of datasets used for their training, we compare the trait distributions of the language models and their underlying corpora. Specifically, we evaluate the traits of \textsc{WebText Test} dataset and text responses generated using GPT-2. Our findings, as shown in Figure~\ref{fig:dist_comp}, indicate that the trait distributions of the \textsc{WebText Test} dataset and randomly generated text using GPT-2, without taking into account any specific type of input prompts, are nearly identical. This suggests that GPT-2 inherits the traits of its training dataset. However, when we passed the test from the Big Five questionnaire as input prompts, we noticed a significant difference in the trait distribution of the \textsc{WebText Test} dataset compared to GPT-2 as measured using the Wasserstein distance (see Table~\ref{tab:wd}). It is important to note that relying solely on random generation may not accurately represent the personality traits of the GPT-2 model. This limitation arises due to the training of GPT-2 on WebText data, which primarily consists of factual information lacking specific personality characteristics. Consequently, randomly generated text may not reflect the language model's inherent traits. To address this limitation, we prompted the language model using the Big Five questionnaire, which allowed us to drive the generated text to reflect the traits indicative of the language model. Using the questionnaire as a guide enables an accurate reflection of the personality traits of the language model, providing a more nuanced and reliable picture of its traits.

\begin{table}[]
\scriptsize
\centering
\begin{tabular}{|c|c|c|}
\hline
\textbf{Traits} &
  \textbf{\begin{tabular}[c]{@{}c@{}}Wasserstein distance\\ dataset vs GPT-2 with prompt\end{tabular}} &
  \textbf{\begin{tabular}[c]{@{}c@{}}Wasserstein distance\\ dataset vs GPT-2 no prompt\end{tabular}} \\ \hline
Agreeableness       & 0.205 & 0.102 \\ \hline
Conscientiousness   & 0.427 & 0.077 \\ \hline
Extraversion        & 0.630 & 0.048 \\ \hline
Emotional Stability & 0.154 & 0.032 \\ \hline
Openness            & 0.139 & 0.025 \\ \hline
\end{tabular}
\caption{Wasserstein distance between trait distributions of WebText dataset and GPT-2 model evaluated with and without input prompts from the Big Five Questionnaire.}
\label{tab:wd}
\end{table}

\subsection{Altering the Traits of Models}

\paragraph{Setup:}
To investigate the methods for altering traits, we restrict ourselves to working with GPT-2 due to limitations on the computational resources available. 

To evaluate \emph{Method 1} discussed in Section~\ref{modify_trait}, we filter the~\cite{siop:big5} dataset by retaining text responses labeled with Big Five factor scores greater than $4$ for the individual traits. We finetune the GPT-2 model on the filtered dataset corresponding to each trait independently and subsequently evaluate the generated text responses. For finetuning, we set the batch size to $16$ and the number of epochs to $20$ with warmup proportion set to $0$, learning rate set to $1e^{-5}$, and weight decay set to $0.01$. 

Similar to \emph{Method 1}, we analyze \emph{Method 2} using the same filtered dataset \cite{siop:big5} according to the following criteria. The personality trait scores in our dataset are continuous values from $1$ to $5$; To set up the model for binary classification, we derive binary target labels for different thresholds from the set $\{2.5, 3, 3.5, 4, 4.5\}$. To illustrate, for a given threshold of 2.5, all the text responses annotated with a score less than 2.5 are labeled as zero and vice-versa. We finetune the original model using this classification task. In particular, we use the standard cross-entropy loss and the Adam optimizer~\citep{kingma2014adam}. We set the learning rate to $5e^{-5}$ and the number of epochs to $10$ while finetuning.

\begin{figure}[htbp]
\captionsetup[subfigure]{font=scriptsize,labelfont=scriptsize}
\begin{multicols}{3}
    \subcaptionbox{Original}{\includegraphics[width=\linewidth]{gpt2.png}}    \par
    \subcaptionbox{Agreeableness}{\includegraphics[width=\linewidth]{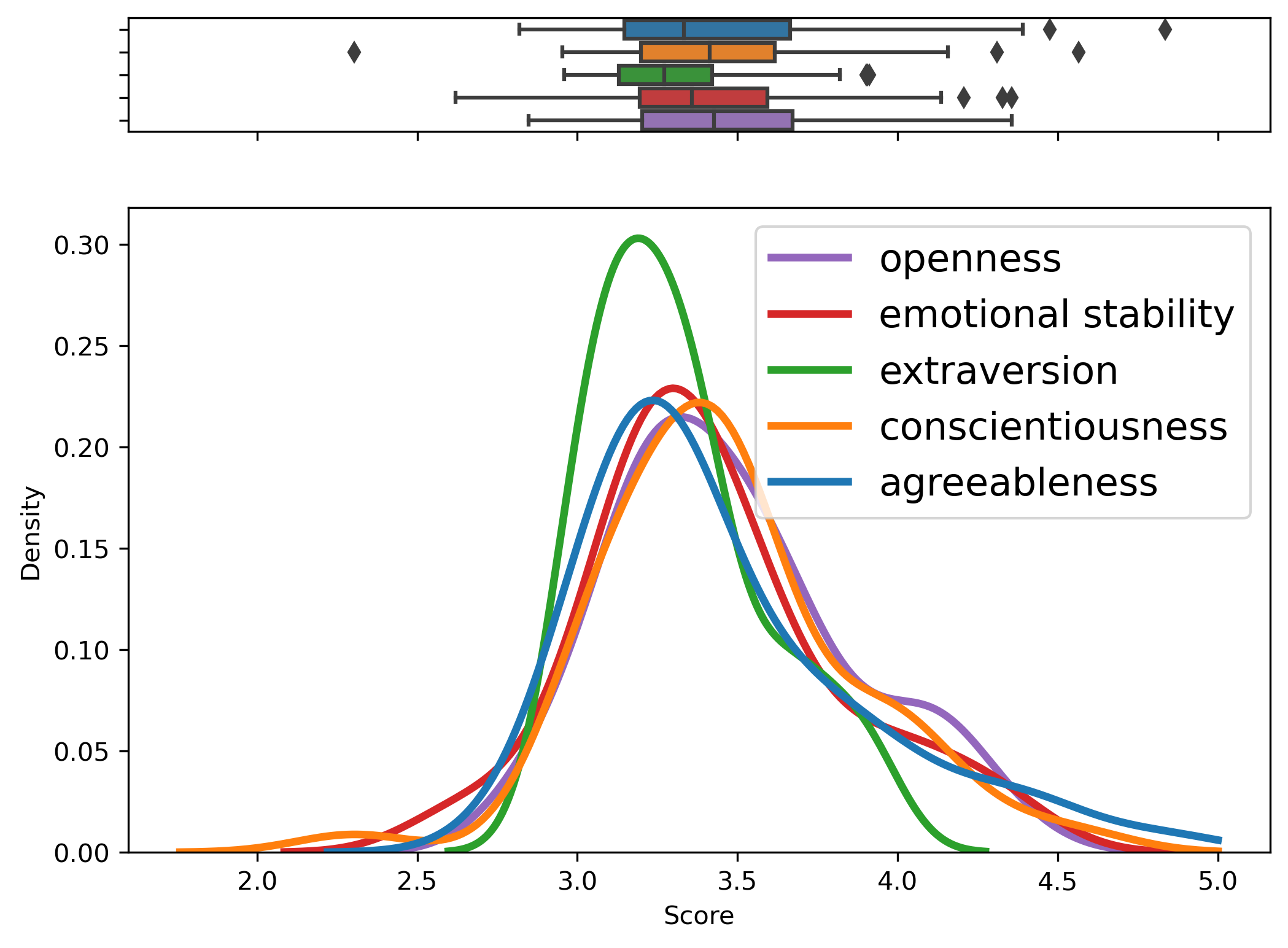}}\par
    \subcaptionbox{Conscientiousness}{\includegraphics[width=\linewidth]{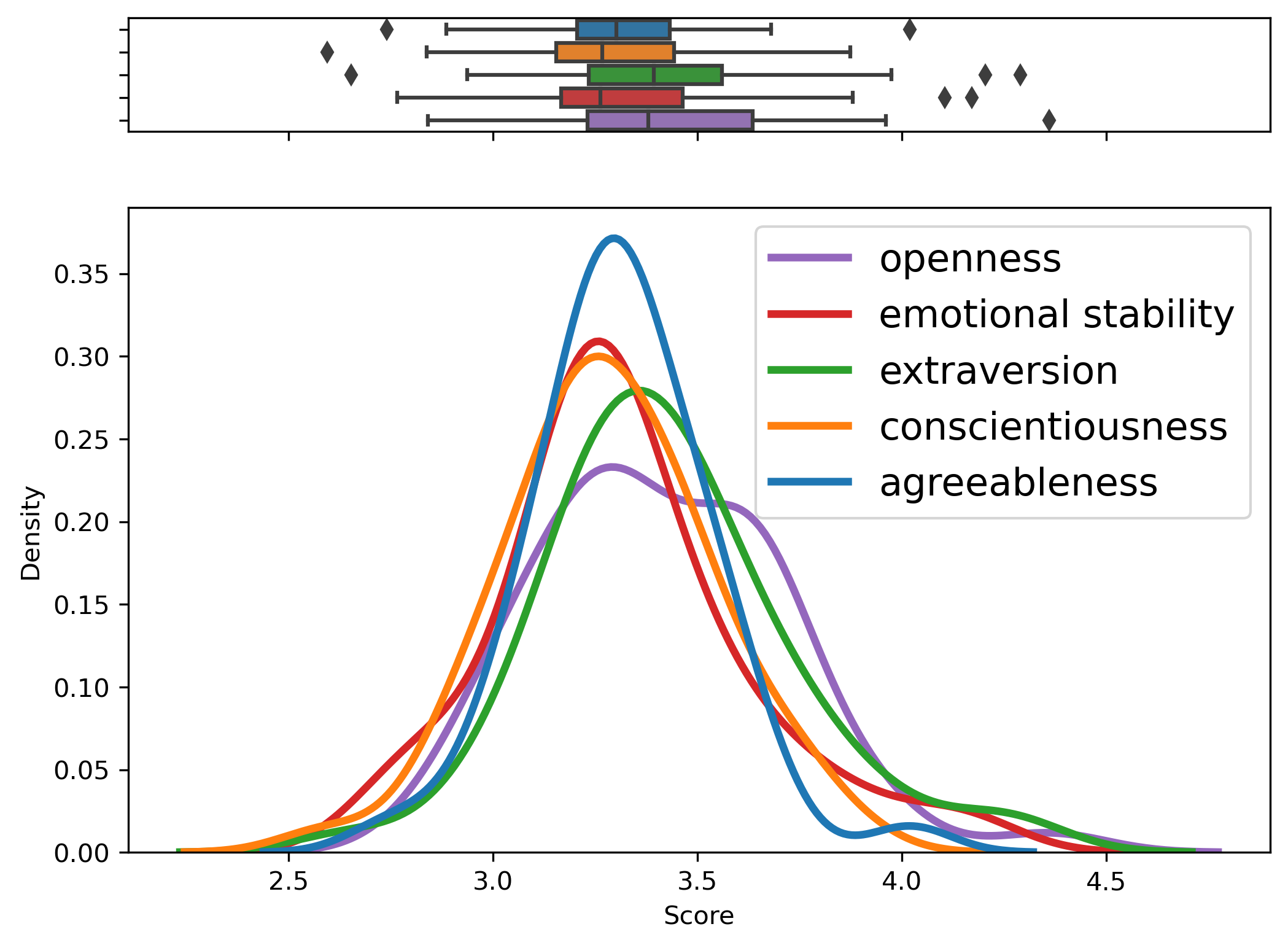}}\par
\end{multicols}
\begin{multicols}{3}
    \subcaptionbox{Extraversion}{\includegraphics[width=\linewidth]{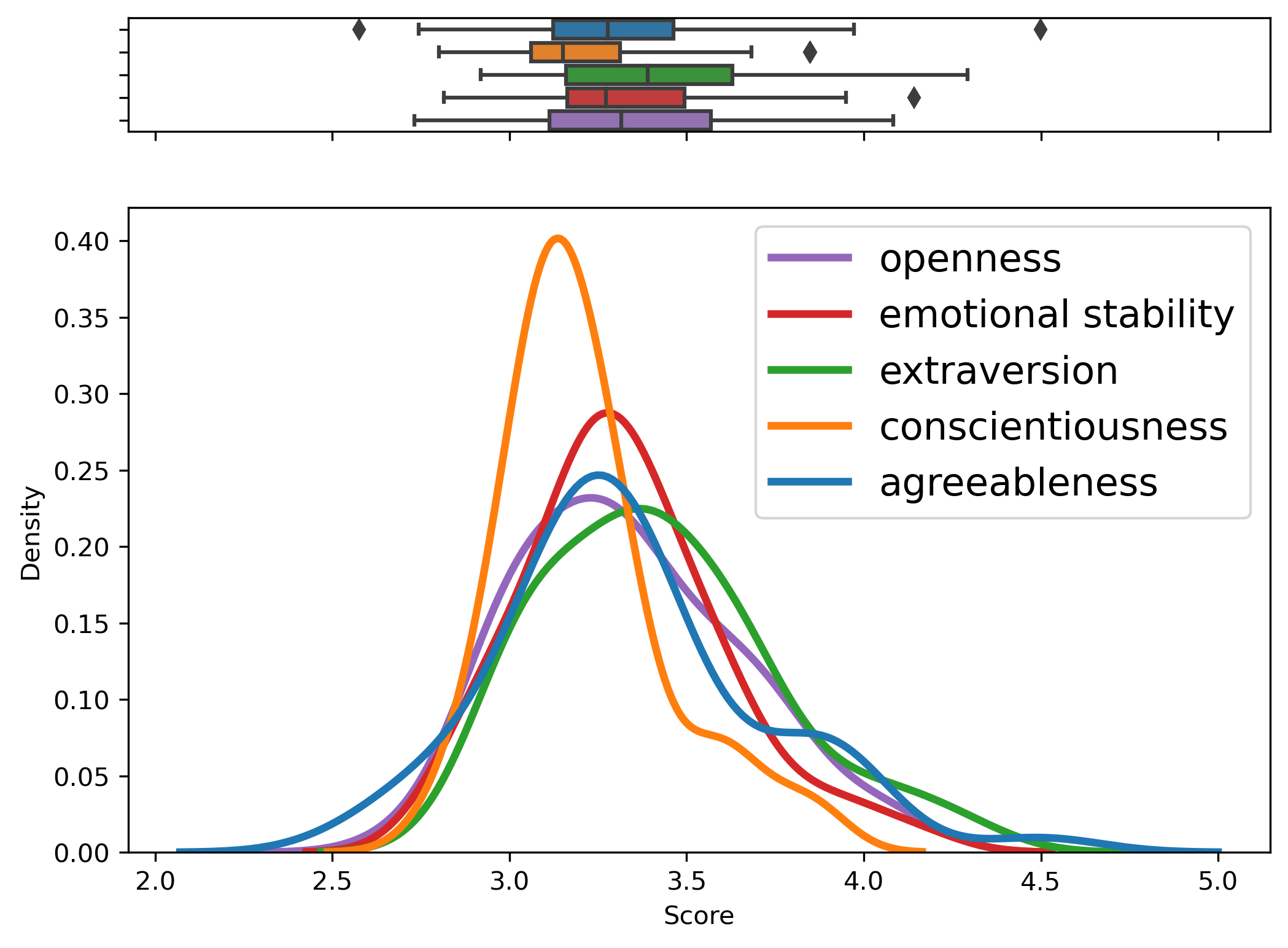}}\par
    \subcaptionbox{Emotional Stability}{\includegraphics[width=\linewidth]{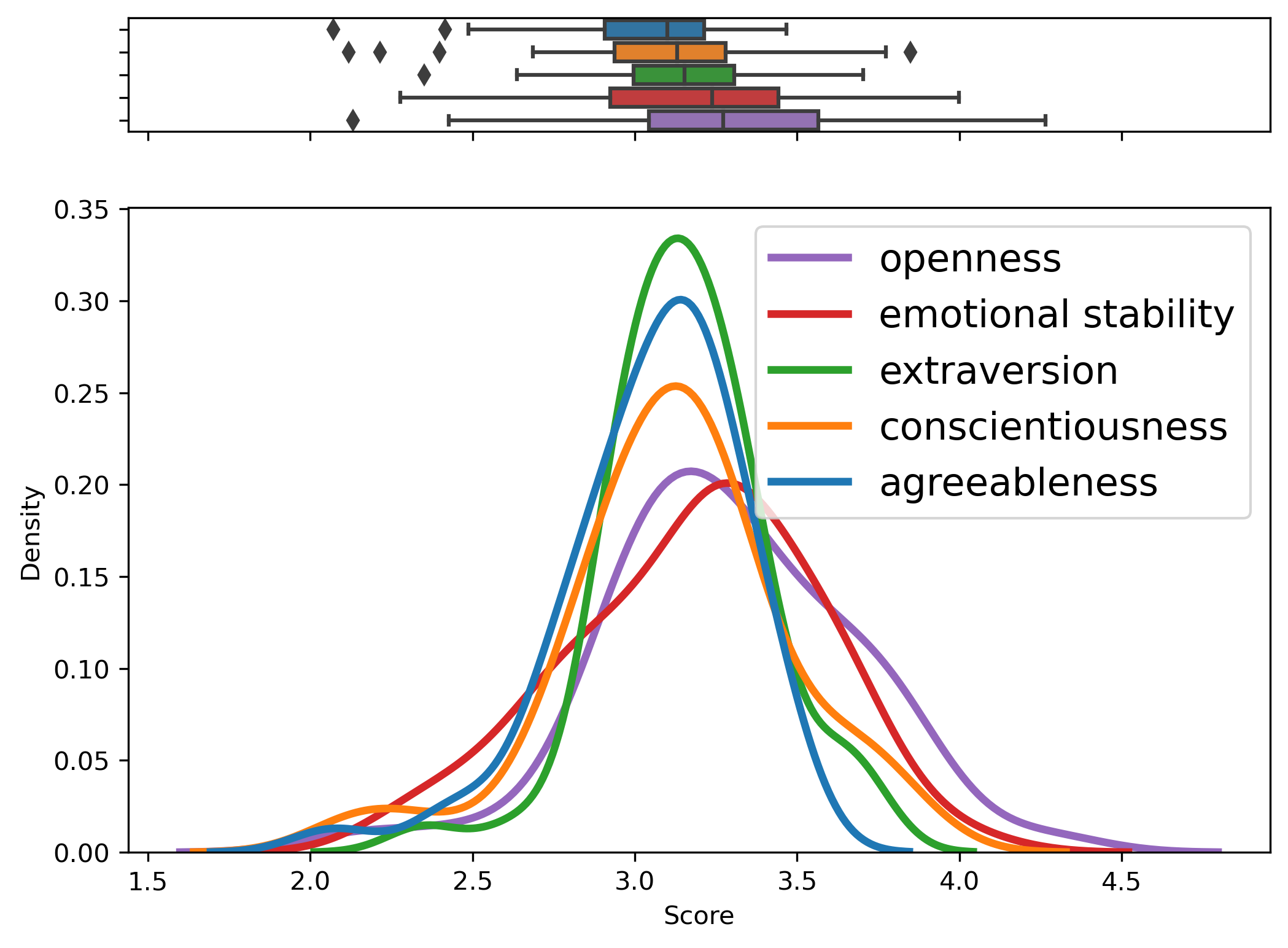}}\par
    \subcaptionbox{Openness}{\includegraphics[width=\linewidth]{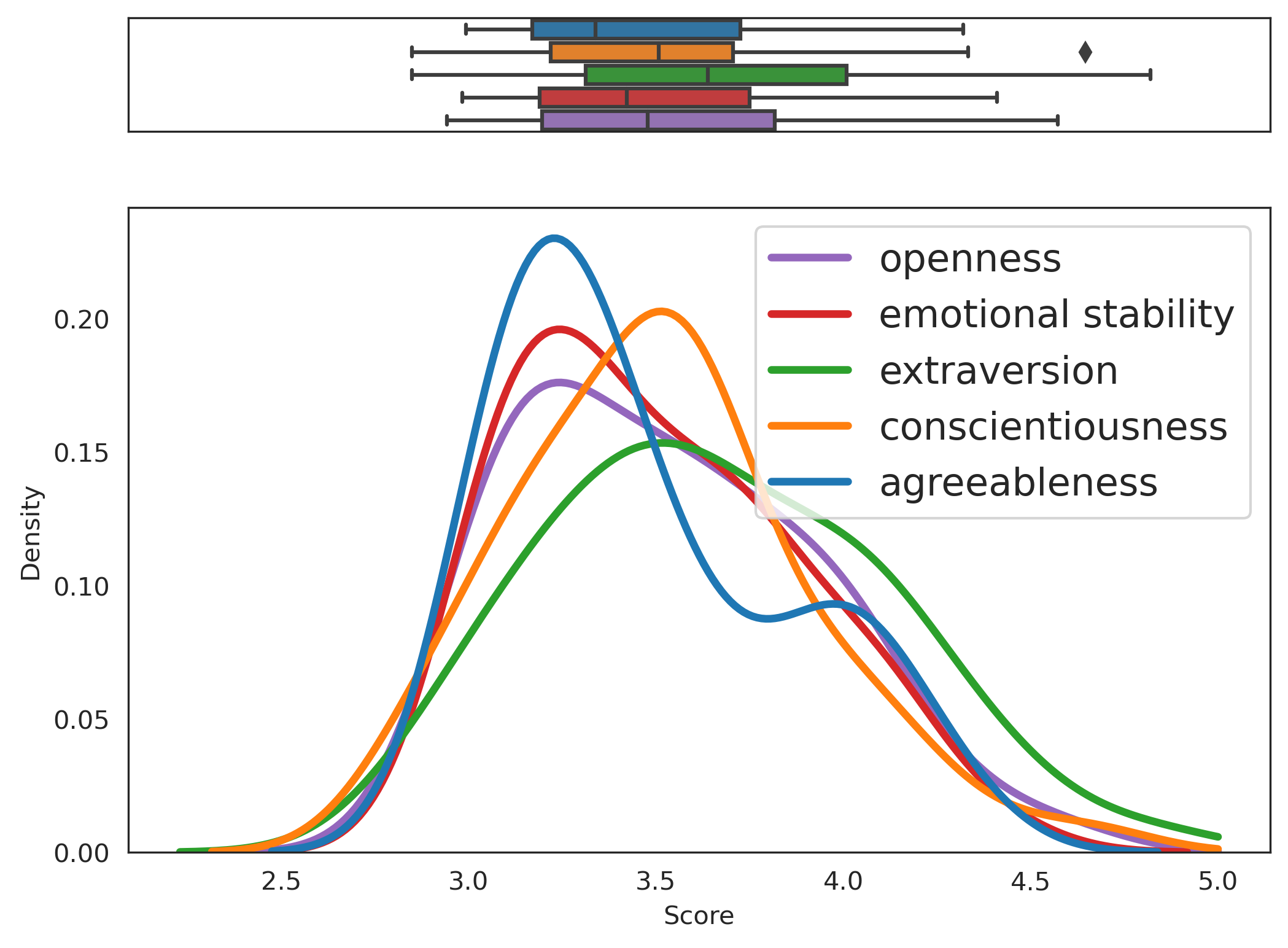}}\par   
\end{multicols}
\caption{Personality trait distributions of finetuned GPT-2.}
\label{ft_dist}
\end{figure}

\paragraph{Results:}
Table \ref{tab:finetune_table} presents the results of \emph{Method 1} for finetuning GPT-2, which indicate a significant change in the personality scores for \emph{Agreeableness}, \emph{Conscientiousness}, \emph{Emotional Stability}, and \emph{Openness} after finetuning with the filtered datasets. These changes are also reflected in the personality trait distributions of the finetuned GPT-2, as shown in Figure \ref{ft_dist}. Our findings demonstrate that finetuning can alter the personality traits of language models like GPT-2 in an open-loop setting by allowing them to learn from a new data corpus. However, we also note that finetuning can change the personality scores of other traits besides the focal trait represented by the filtered dataset, which may be undesirable if precise control over a specific trait improvement is required. This issue remains an avenue for future research.

\begin{table}[htbp]
\centering
\resizebox{\textwidth}{!}{%
\begin{tabular}{|c|c|c|c|c|c|c|}
\hline
Trait               & Before finetuning & Agreeableness        & Conscientiousness & Extraversion         & Emotional Stability  & Openness             \\ \hline
Agreeableness       & 3.41 (0.73)       & 3.33 (0.43) & 3.30 (0.21)       & 3.27 (0.36)          & 3.09 (0.27)          & 3.33 (0.37)          \\ \hline
Conscientiousness   & 3.18 (0.39)       & 3.41 (0.39)          & \textbf{3.26 (0.25)}       & 3.15 (0.22)          & 3.13 (0.34)          & 3.50 (0.38)          \\ \hline
Extraversion        & 3.07 (0.60)       & 3.27 (0.26)          & 3.39 (0.30)       & \textbf{3.39 (0.32)} & 3.15 (0.24)          & 3.63 (0.44)          \\ \hline
Emotional Stability & 3.15 (0.50)       & 3.35 (0.38)          & 3.26 (0.29)       & 3.27 (0.28)          & \textbf{3.23 (0.37)} & 3.42 (0.36)          \\ \hline
Openness            & 2.97 (0.47)       & 3.42 (0.36)          & 3.37 (0.29)       & 3.31 (0.31)          & 3.27 (0.38)          & \textbf{3.47 (0.39)} \\ \hline
\end{tabular}%
}
\caption{Personality scores with their uncertainties of the finetuned GPT-2.}
\label{tab:finetune_table}
\end{table}

Table \ref{tab:Extraversion-table} compares median scores for personality traits of the GPT-2 model finetuned using \emph{Method 2} on \emph {Extraversion} annotated labels at different thresholds. We observe improvement \emph{Extraversion} scores of the finetuned GPT-2 models compared to the original GPT-2 model. However, we also observe noticeable changes in the scores of other personality traits, similar to \emph{Method 1}, which is not desirable. We noticed similar changes in personality trait scores when the model was finetuned using other personality-annotated datasets, as shown in~\ref{appendix}.
\begin{table}[htbp]
\centering
\resizebox{\textwidth}{!}{%
\begin{tabular}{|c|c|c|c|c|c|c|}
\hline
Trait &
  Before finetuning &
  \begin{tabular}[c]{@{}c@{}}After finetuning \\ at threshold (2.5)\end{tabular} &
  \begin{tabular}[c]{@{}c@{}}After finetuning \\ at threshold (3.0)\end{tabular} &
  \begin{tabular}[c]{@{}c@{}}After finetuning\\ at threshold (3.5)\end{tabular} &
  \begin{tabular}[c]{@{}c@{}}After finetuning \\ at threshold (4.0)\end{tabular} &
  \begin{tabular}[c]{@{}c@{}}After finetuning \\ at threshold (4.5)\end{tabular} \\ \hline
Agreeableness     & 3.41 (0.73) & 3.02 (0.48) & 3.19 (0.54) & 3.33 (0.64) & 3.14 (0.76) & 3.24 (0.57) \\ \hline
Conscientiousness & 3.18 (0.39) & 3.19 (0.53) & 3.11 (0.49) & 3.17 (0.67) & 3.04 (0.57) & 3.19 (0.57) \\ \hline
\textit{Extraversion}      & \textit{3.07 (0.60)} & \textit{2.89 (0.51)} & \textit{3.19 (0.53)} & \textit{3.24 (0.75)} & \textit{2.90 (0.74)} & \textit{3.25 (0.49)} \\ \hline
Emotional Stability      & 3.15 (0.50) & 3.12 (0.67) & 2.99 (0.45) & 3.19 (0.64) & 3.00 (0.74) & 3.11 (0.46) \\ \hline
Openness          & 2.97 (0.47) & 2.96 (0.51) & 3.09 (0.55) & 3.13 (0.67) & 2.88 (0.80) & 3.15 (0.54) \\ \hline
\end{tabular}%
}
\caption{Personality scores before and after finetuning the GPT-2 using \emph{Extraversion} labeled data at different thresholds using the \emph{Method 2}.}
\label{tab:Extraversion-table}
\end{table}

Overall, our methods for altering personality traits show initial promise in altering the personality traits of language models. But some of the challenges still remain. Specifically, since the personality of a language model is closely tied to its text generation capabilities, changing one of them may affect the other. Further work in this direction would be a valuable endeavor.

\section{Conclusion} \label{sec:conclusion}
In this paper, we developed approaches to quantify the personality traits of datasets and language models designed for open-ended text generation. We presented a principled approach for evaluating the personality traits of datasets using a pre-trained zero-shot classifier. We then extended this idea to estimate personality profiles of language models using the prompts from the Big Five personality test questionnaire. Our experiments revealed that language models possess varied personality traits reflecting the datasets used in their training, thus influencing their generated outputs. Furthermore, we proposed straightforward approaches to alter the personality traits of white-box language models by finetuning them on personality-annotated datasets. Overall, our work provides a critical starting point for assessing and understanding the personality traits of language models.

In future research, we aim to design robust approaches that can precisely alter the personality traits of language models. We also aim to integrate other commonly used assessments like the Myer Briggs Type indicator in assessing, validating, and modifying the personality traits of language models.

\bibliography{references}
\appendix
\section{Additional Details} \label{appendix}
\paragraph{Wasserstein Distance ($d_p$)}
It is a measure of the distance between two probability distributions. It quantifies the amount of work required to transform one distribution into the other. The Wasserstein distance between two cumulative distributions, P and Q, is given by:

\begin{equation}
    d_p(P, Q) := \underset{U, V}{inf}||U-P||_p
\end{equation}

Where the infimum is taken over all pairs of random variables $(U, V)$ with respective cumulative distributions P and Q.

\paragraph{Adam}
It is an optimization algorithm that extends the capabilities of stochastic gradient descent (SGD) by incorporating adaptive learning rates for all the parameters. The adaptive learning rates allow the optimizer to converge faster and more accurately, even in noisy or sparse gradients.

\paragraph{Learning Rate} It is a hyperparameter that controls the amount by which the model weights need to be updated with respect to gradient loss during the training. 

\paragraph{Epoch} It refers to one complete pass through the entire training dataset during the training of a model

\paragraph{Weight Decay} It is a regularization hyperparameter used to prevent the overfitting of models. It works by adding a penalty term to the loss function during training, which enables the model to learn simpler and smoother weight configurations.

\paragraph{Warmup Proportion} It is a hyperparameter that determines the percentage of training steps during which the learning rate is gradually increased from an initial value to its maximum value
\\~\\
Tables \ref{tab:Conscientiousness-table}, \ref{tab:Openness-table}, \ref{tab:Emotionalstability-table} and \ref{tab:Agreeableness-table}  shows detailed results of finetuning GPT-2 model using \emph{Method 2} on personality annotated data.

\begin{table}[htbp]
\centering
\resizebox{\textwidth}{!}{%
\begin{tabular}{|c|c|c|c|c|c|c|}
\hline
Trait &
  Before finetuning &
  \begin{tabular}[c]{@{}c@{}}After finetuning \\ at threshold (2.5)\end{tabular} &
  \begin{tabular}[c]{@{}c@{}}After finetuning \\ at threshold (3.0)\end{tabular} &
  \begin{tabular}[c]{@{}c@{}}After finetuning\\ at threshold (3.5)\end{tabular} &
  \begin{tabular}[c]{@{}c@{}}After finetuning \\ at threshold (4.0)\end{tabular} &
  \begin{tabular}[c]{@{}c@{}}After finetuning \\ at threshold (4.5)\end{tabular} \\ \hline
Agreeableness     & 3.41 (0.73) & 3.51 (0.51) & 3.19 (0.30) & 3.19 (0.44) & 3.10 (0.68) & 3.04 (0.45) \\ \hline
Conscientiousness & 3.18 (0.39) & 3.29 (0.53) & 3.17 (0.41) & 3.13 (0.34) & 3.26 (0.57) & 3.36 (0.47) \\ \hline
Extraversion      & 3.07 (0.60) & 3.24 (0.55) & 3.13 (0.29) & 3.20 (0.27) & 3.00 (0.63) & 3.04 (0.42) \\ \hline
Emotional Stability      & 3.15 (0.46) & 3.42 (0.53) & 3.15 (0.37) & 3.11 (0.42) & 3.09 (0.70) & 2.94 (0.34) \\ \hline
Openness          & 2.97 (0.47) & 3.33 (0.56) & 3.23 (0.36) & 3.26 (0.38) & 3.03 (0.50) & 3.11 (0.45) \\ \hline
\end{tabular}%
}
\vspace{3mm}
\caption{Personality scores before and after finetuning the GPT-2 using \emph{Conscientiousness} labeled data at different thresholds using the \emph{Method 2}}
\label{tab:Conscientiousness-table}
\end{table}

\begin{table}[htbp]
\centering
\resizebox{\textwidth}{!}{%
\begin{tabular}{|c|c|c|c|c|c|c|}
\hline
Trait &
  Before finetuning &
  \begin{tabular}[c]{@{}c@{}}After finetuning \\ at threshold (2.5)\end{tabular} &
  \begin{tabular}[c]{@{}c@{}}After finetuning \\ at threshold (3.0)\end{tabular} &
  \begin{tabular}[c]{@{}c@{}}After finetuning\\ at threshold (3.5)\end{tabular} &
  \begin{tabular}[c]{@{}c@{}}After finetuning \\ at threshold (4.0)\end{tabular} &
  \begin{tabular}[c]{@{}c@{}}After finetuning \\ at threshold (4.5)\end{tabular} \\ \hline
Agreeableness     & 3.41 (0.73) & 3.05 (0.45) & 3.22 (0.34) & 2.83 (0.49) & 2.99 (0.71) & 3.06 (0.13) \\ \hline
Conscientiousness & 3.18 (0.39) & 3.08 (0.38) & 3.28 (0.49) & 2.93 (0.49) & 3.12 (0.72) & 3.09 (0.11) \\ \hline
Extraversion      & 3.07 (0.60) & 3.03 (0.41) & 3.30 (0.46) & 2.90 (0.47) & 3.11 (0.71) & 3.08 (0.10) \\ \hline
Emotional Stability       & 3.15 (0.50) & 3.09 (0.37) & 3.17 (0.41) & 2.93 (0.45) & 3.09 (0.60) & 3.08 (0.14) \\ \hline
Openness          & 2.97 (0.47) & 3.08 (0.38) & 3.22 (0.44) & 3.07 (0.32) & 3.04 (0.78) & 3.09 (0.12) \\ \hline
\end{tabular}%
}
\vspace{3mm}
\caption{Personality scores before and after finetuning the GPT-2 using \emph{Openness} labeled data at different thresholds using the \emph{Method 2}}
\label{tab:Openness-table}
\end{table}

\begin{table}[htbp]
\centering
\resizebox{\textwidth}{!}{%
\begin{tabular}{|c|c|c|c|c|c|c|}
\hline
Trait &
  Before finetuning &
  \begin{tabular}[c]{@{}c@{}}After finetuning \\ at threshold (2.5)\end{tabular} &
  \begin{tabular}[c]{@{}c@{}}After finetuning \\ at threshold (3.0)\end{tabular} &
  \begin{tabular}[c]{@{}c@{}}After finetuning\\ at threshold (3.5)\end{tabular} &
  \begin{tabular}[c]{@{}c@{}}After finetuning \\ at threshold (4.0)\end{tabular} &
  \begin{tabular}[c]{@{}c@{}}After finetuning \\ at threshold (4.5)\end{tabular} \\ \hline
Agreeableness     & 3.41 (0.73) & 3.83 (0.55) & 3.05 (0.49) & 2.89 (0.18) & 2.93 (0.73) & 3.04 (0.69) \\ \hline
Conscientiousness & 3.18 (0.39) & 3.31 (0.55) & 2.98 (0.42) & 2.90 (0.15) & 3.02 (0.80) & 3.05 (0.80) \\ \hline
Extraversion      & 3.07 (0.60) & 3.38 (0.77) & 3.06 (0.47) & 2.90 (0.13) & 2.99 (0.76) & 3.20 (0.58) \\ \hline
Emotional Stability       & 3.15 (0.50) & 3.16 (0.52) & 3.01 (0.55) & 3.17 (0.12) & 3.14 (0.75) & 3.22 (0.71) \\ \hline
Openness          & 2.97 (0.47) & 3.15 (0.59) & 3.19 (0.51) & 2.91 (0.17) & 3.19 (0.76) & 3.07 (0.83) \\ \hline
\end{tabular}%
}
\vspace{3mm}
\caption{Personality scores before and after finetuning the GPT-2 using \emph{Emotional stability} labeled data at different thresholds using the \emph{Method 2}}
\label{tab:Emotionalstability-table}
\end{table}

\begin{table}[htbp]
\centering
\resizebox{\textwidth}{!}{%
\begin{tabular}{|c|c|c|c|c|c|c|}
\hline
Trait &
  Before finetuning &
  \begin{tabular}[c]{@{}c@{}}After finetuning \\ at threshold (2.5)\end{tabular} &
  \begin{tabular}[c]{@{}c@{}}After finetuning \\ at threshold (3.0)\end{tabular} &
  \begin{tabular}[c]{@{}c@{}}After finetuning\\ at threshold (3.5)\end{tabular} &
  \begin{tabular}[c]{@{}c@{}}After finetuning \\ at threshold (4.0)\end{tabular} &
  \begin{tabular}[c]{@{}c@{}}After finetuning \\ at threshold (4.5)\end{tabular} \\ \hline
Agreeableness     & 3.41 (0.73) & 3.39 (0.73) & 3.37 (0.63) & 3.42 (0.51) & 3.49 (0.23) & 3.55 (0.36) \\ \hline
Conscientiousness & 3.18 (0.39) & 3.10 (0.80) & 2.88 (0.57) & 3.03 (0.70) & 3.06 (0.31) & 3.04 (0.24) \\ \hline
Extraversion      & 3.07 (0.60) & 3.06 (0.66) & 2.98 (0.61) & 3.01 (0.40) & 3.05 (0.24) & 3.08 (0.38) \\ \hline
Emotional Stability       & 3.15 (0.50) & 3.08 (0.66) & 3.00 (0.65) & 3.04 (0.40) & 3.03 (0.19) & 3.03 (0.25) \\ \hline
Openness          & 2.97 (0.47) & 3.28 (0.66) & 2.91 (0.62) & 3.16 (0.32) & 3.04 (0.13) & 3.06 (0.22) \\ \hline
\end{tabular}%
}
\vspace{3mm}
\caption{Personality scores before and after finetuning the GPT-2 using \emph{Agreeableness} labeled data at different thresholds using the \emph{Method 2}}
\label{tab:Agreeableness-table}
\end{table}

\begin{table}[]

\label{tab:questions}
\resizebox{\textwidth}{!}{%
\begin{tabular}{|l|l|}
\hline
I am the life of the party.                     & I have little to say.                                    \\ \hline
I feel little concern for others.               & I have a soft heart.                                     \\ \hline
I am always prepared.                           & I often forget to put things back in their proper place. \\ \hline
I get stressed out easily.                      & I get upset easily.                                      \\ \hline
I have a rich vocabulary.                       & I do not have a good imagination.                        \\ \hline
I don't talk a lot.                             & I talk to a lot of different people at parties.          \\ \hline
I am interested in people.                      & I am not really interested in others.                    \\ \hline
I leave my belongings around.                   & I like order.                                            \\ \hline
I am relaxed most of the time.                  & I change my mood a lot.                                  \\ \hline
I have difficulty understanding abstract ideas. & I am quick to understand things.                         \\ \hline
I feel comfortable around people.               & I don't like to draw attention to myself.                \\ \hline
I insult people.                                & I take time out for others.                              \\ \hline
I pay attention to details.                     & I shirk my duties.                                       \\ \hline
I worry about things.                           & I have frequent mood swings.                             \\ \hline
I have a vivid imagination.                     & I use difficult words.                                   \\ \hline
I keep in the background.                       & I don't mind being the center of attention.              \\ \hline
I sympathize with others' feelings.             & I feel others' emotions.                                 \\ \hline
I make a mess of things.                        & I follow a schedule.                                     \\ \hline
I seldom feel blue.                             & I get irritated easily.                                  \\ \hline
I am not interested in abstract ideas.          & I spend time reflecting on things.                       \\ \hline
I start conversations.                          & I am quiet around strangers.                             \\ \hline
I am not interested in other people's problems. & I make people feel at ease.                              \\ \hline
I get chores done right away.                   & I am exacting in my work.                                \\ \hline
I am easily disturbed.                          & I often feel blue.                                       \\ \hline
I have excellent ideas.                         & I am full of ideas.                                      \\ \hline
\end{tabular}%

}
\vspace{3mm}
\caption{Questions from the Big Five questionnaire}
\end{table}

\end{document}